\documentclass[cameraready]{Interspeech}
\usepackage{amsmath,graphicx,hyperref}
\usepackage{tikz}
\usetikzlibrary{shapes, positioning}
\usepackage{float}
\usepackage{hyperref}
\usepackage{cite}
\usepackage{subcaption}
\usepackage{multirow}
\usepackage{booktabs}
\title{Investigating Faithfulness in Large Audio Language Models}

\author[affiliation={1, 2}, correspondingauthor]{Pooneh}{Mousavi}
\author[affiliation={4}]{Lovenya}{Jain}
\author[affiliation={1, 2}]{Mirco}{Ravanelli}
\author[affiliation={3, 2}]{Cem}{Subakan}

\address{
    $^1$ Concordia University, Canada, 
    $^2$ Mila - Quebec AI Institute, Canada \\
    $^3$ Université Laval, Canada ,
     $^4$ Birla Institute of Technology and Science, Pilani 
}
\email{pooneh.mousavi@mila.quebec}


\usepackage{comment}

\begin{document}

\maketitle

\begin{tikzpicture}[remember picture,overlay]
    \node[rotate=90, text=gray, font=\small] 
    at ([xshift=1.5cm]current page.west) 
    {Accepted at INTERSPEECH 2026};
\end{tikzpicture}

\begin{abstract}

Large Audio Language Models (LALMs) integrate audio encoders with pretrained Large Language Models to perform complex multimodal reasoning tasks. While these models can generate Chain-of-Thought (CoT) explanations, the faithfulness of these reasoning chains remains unclear. In this work, we propose a systematic framework to evaluate CoT faithfulness in LALMs with respect to both the input audio and the final model prediction. We define three criteria for audio faithfulness: hallucination-free, holistic, and attentive listening. We also introduce a benchmark based on both audio and CoT interventions to assess faithfulness\footnote{The benchmarking interface and evaluation results are available at \url{https://poonehmousavi.github.io/faithfulness/}.}. Experiments on Audio Flamingo 3 and Qwen2.5-Omni suggest a potential multimodal disconnect: reasoning often aligns with the final prediction but is not always strongly grounded in the audio and can be vulnerable to hallucinations or adversarial perturbations. 

%Multimodal Large Language Models come with the capability of providing Chain-of-Thought (CoT) representations that would help to explain their reasoning for their decision making. The concept of faithfulness aims to measure whether these CoTs accurately reflect the model's decision process, and therefore can be used as reliable explanations. 

%In this paper, we study the faithfulness of CoTs similar to the way it had been studied for text LLMs through introducing interventions on CoTs, to understand how sensitive the underlying model to the produced CoTs. We also take the intervention based faithfulness analysis a step further through audio domain input interventions, with the goal to provide a more complete understanding as to if the model in fact also listens to the audio input or not. Our experimental results with the CoT interventions point towards the fact that the CoTs tend to be faithful, whereas the models do not necessarily have a holistic understanding of the inputted audio. 

\end{abstract}

%todo: vertical line in figure -- done
%todo: error bars on CoT consistency plot

\newcommand{\cem}[1]{ {\color{red} $\mathcal C$ #1 }}

\label{sec:intro}

% \begin{figure*}[t]
%     \centering
%     % Left Figure: Audio Intervention (Larger)
%     \begin{subfigure}[m]{0.55\textwidth}
%         \centering
%         \includegraphics[width=\linewidth, valign=m, trim=0.0cm 0.4cm 0.0cm 0.0cm, clip]{figures/pipeline_audio.pdf}
%         \caption{Audio Intervention Module}
%         \label{fig:pipeline_audio_intervention}
%     \end{subfigure}
%     \hspace{1em} % Small space between them
%     % Right Figure: CoT Reasoning (Smaller & Vertically Centered)
%     \begin{subfigure}[m]{0.38\textwidth}
%         \centering
%         \includegraphics[width=\linewidth, valign=m, trim=0.0cm 0.4cm 0.0cm 0.0cm, clip]{figures/pipeline_cot.pdf}
%         \caption{CoT Intervention Module}
%         \label{fig:pipeline_cot_intervention}
%     \end{subfigure}

%     \caption{\textbf{LALM Consistency Evaluation Framework.} .}
%     \label{fig:full_pipeline_centered}
% \end{figure*}

\section{Introduction}

Large Language Models (LLMs) have transformed machine learning in recent years. In the audio and speech domain, Large Audio Language Models (LALMs) followed suit for tackling complex audio understanding tasks such as Audio Question-Answering. Large Audio-Language Models (LALMs) integrate audio encoders with pre-trained decoder-based LLMs, enabling open-ended audio question-answering and free-form response generation~\cite{aroralandscape,xu2025qwen3,Qwen2.5-Omni,goel2025audioflamingo3advancing,wu2025step,mousavi2025listen}. An important feature of text LLMs is that they can be prompted to provide reasoning for their decisions, potentially helping their deployment in decision-critical applications such as healthcare, security, and forensics~\cite{kojima2022large,zhouleast}. Prior studies show that generating intermediate reasoning steps, often called chain-of-thought (CoT) or reasoning chains, can improve explainability and trustworthiness ~\cite{yao2023reactsynergizingreasoningacting,liexplanations,wang2022rationaleaugmentedensembleslanguagemodels,yao2023tree}. 
CoT provides a reasoning chain for complex tasks, and can make the model output more accurate and interpretable.
%CoT provides a reasoning chain for complex tasks into smaller subproblems and allocates more computation to harder questions, which can make predictions more accurate and interpretable.

Similar to LLMs, LALMs also have the capability to generate Chain-of-Thought (CoT) explanations for their predictions. Previous studies evaluate the robustness of LALMs to audio modifications and prompt perturbations~\cite{hou2025evaluating,lopez2025robustness,li2025silence}. However, they do not investigate whether these CoTs faithfully reflect the model’s internal decision process. This raises a key question for trustworthy AI: \textit{How faithful are the chain-of-thought explanations produced by LALMs?} In particular, we ask the following question: \textit{i)How faithful are the CoTs to the input audio? (i.e., to what extent does the model ground its reasoning in the acoustic input?)}
\textit{ii) How faithful are the CoTs to the final model output? (i.e., how does the final prediction respond to changes in the reasoning path? }  

\begin{figure}[t]
    \centering
    \includegraphics[width=\linewidth, trim=0.0cm 0.4cm 0.0cm 0.0cm, clip]{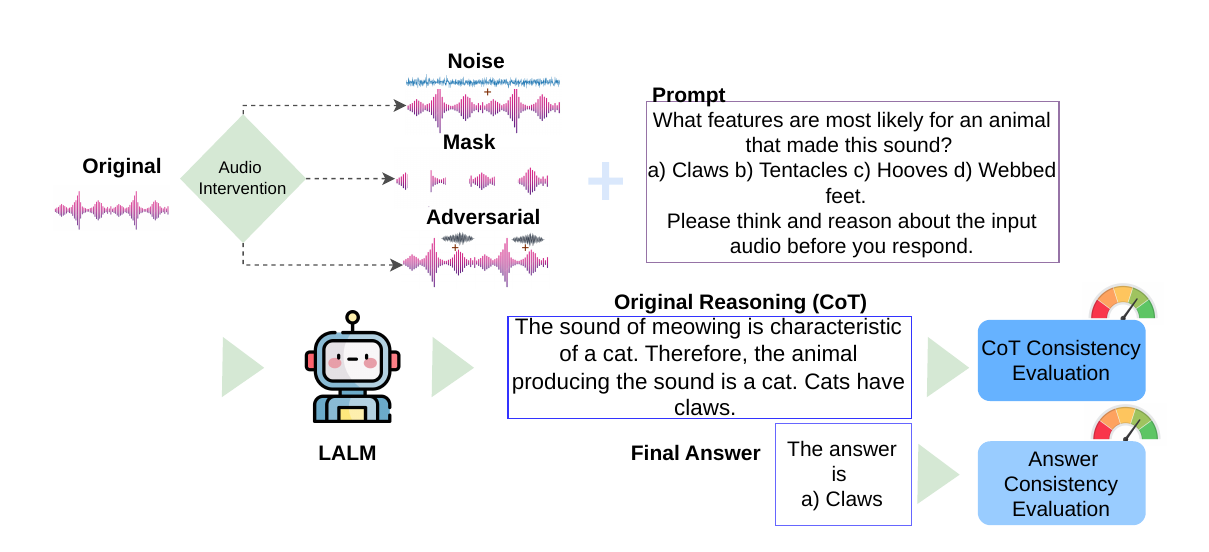}
    \noindent\rule{\linewidth}{0.3mm}
    %\caption{Audio Intervention Pipeline.}
    \includegraphics[width=\linewidth, trim=0.0cm 0.4cm 0.0cm 0.0cm, clip]{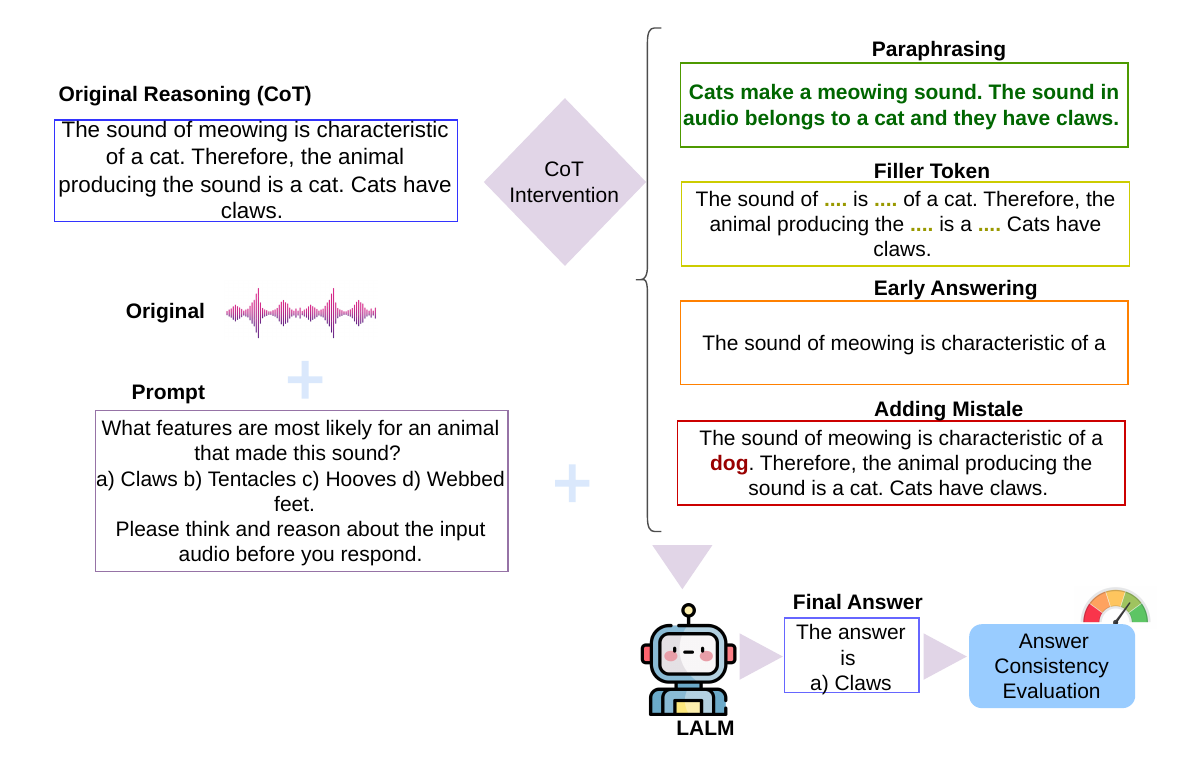}
    \vspace{-.6cm}
    \caption{\textbf{(top)} Audio Intervention Pipeline  \textbf{(bottom)}CoT Intervention Pipeline.}
    \vspace{-.8cm}
    \label{fig:pipeline_cot_intervention}
\end{figure}

\newcommand{\figsizeiii}{0.2}
\begin{figure*}[t]
    \centering
    % Legend Row: Using % to remove whitespace and negative hspace to pull them together
    \includegraphics[width=0.8\linewidth]{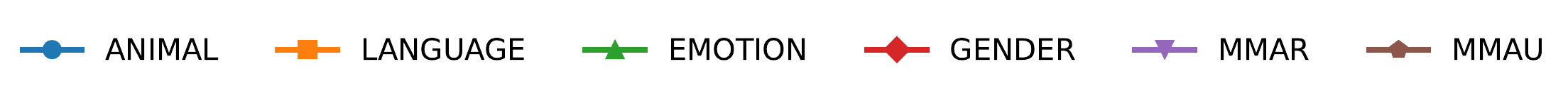}%
    \vspace{-.3cm}
    \hspace{-3mm} 
    \includegraphics[width=0.22\linewidth]{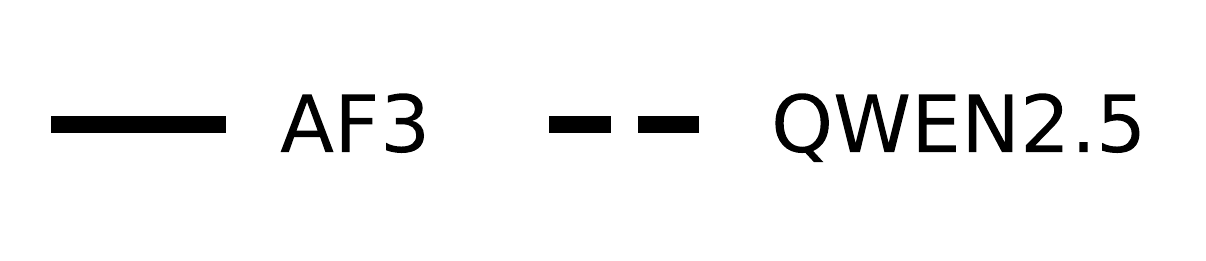}
    
    \vspace{-0.2cm} 

    % Single row of plots with a solid vertical separator
    \includegraphics[width=\figsizeiii\linewidth]{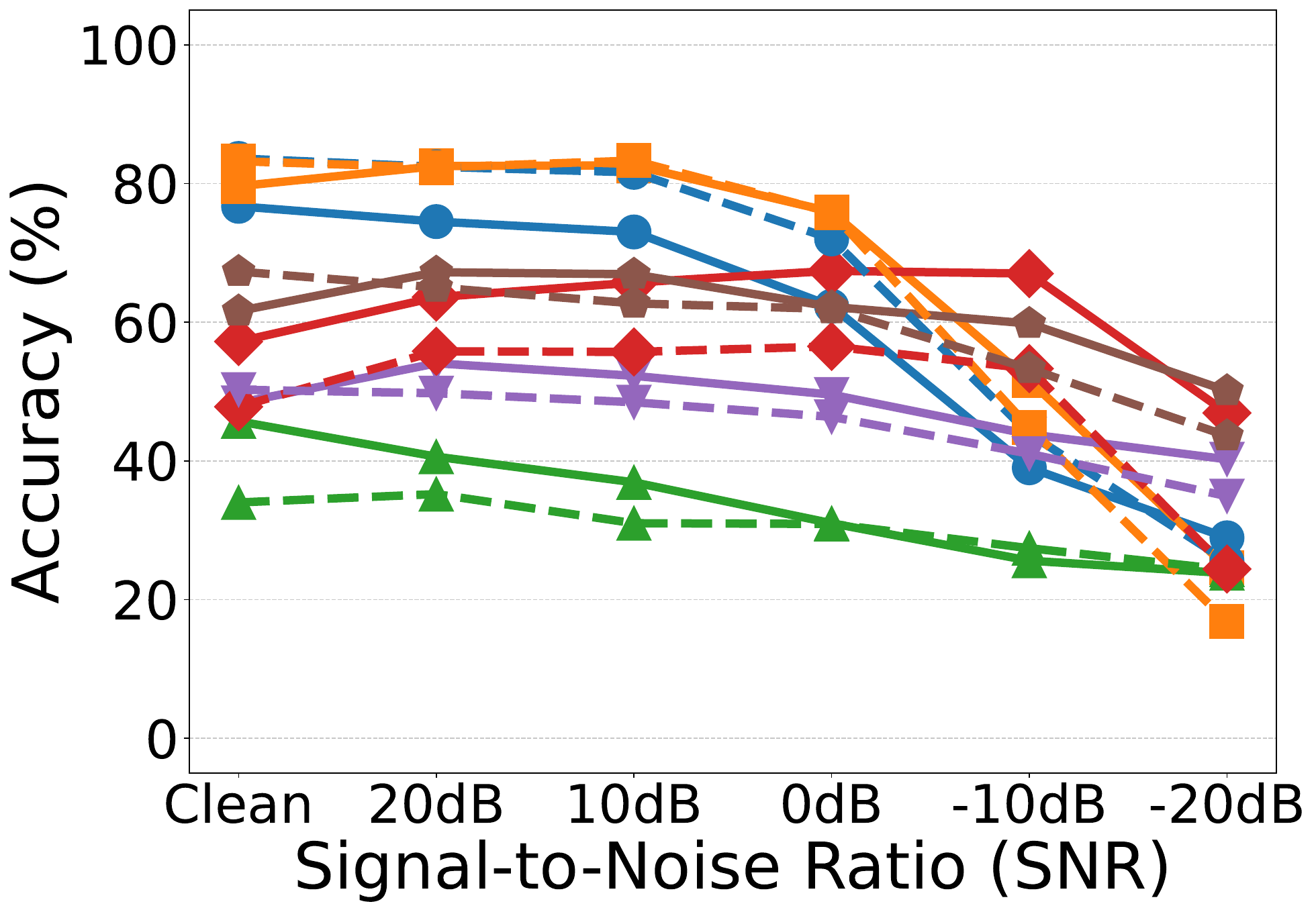}
    \hfill
    \includegraphics[width=\figsizeiii\linewidth]{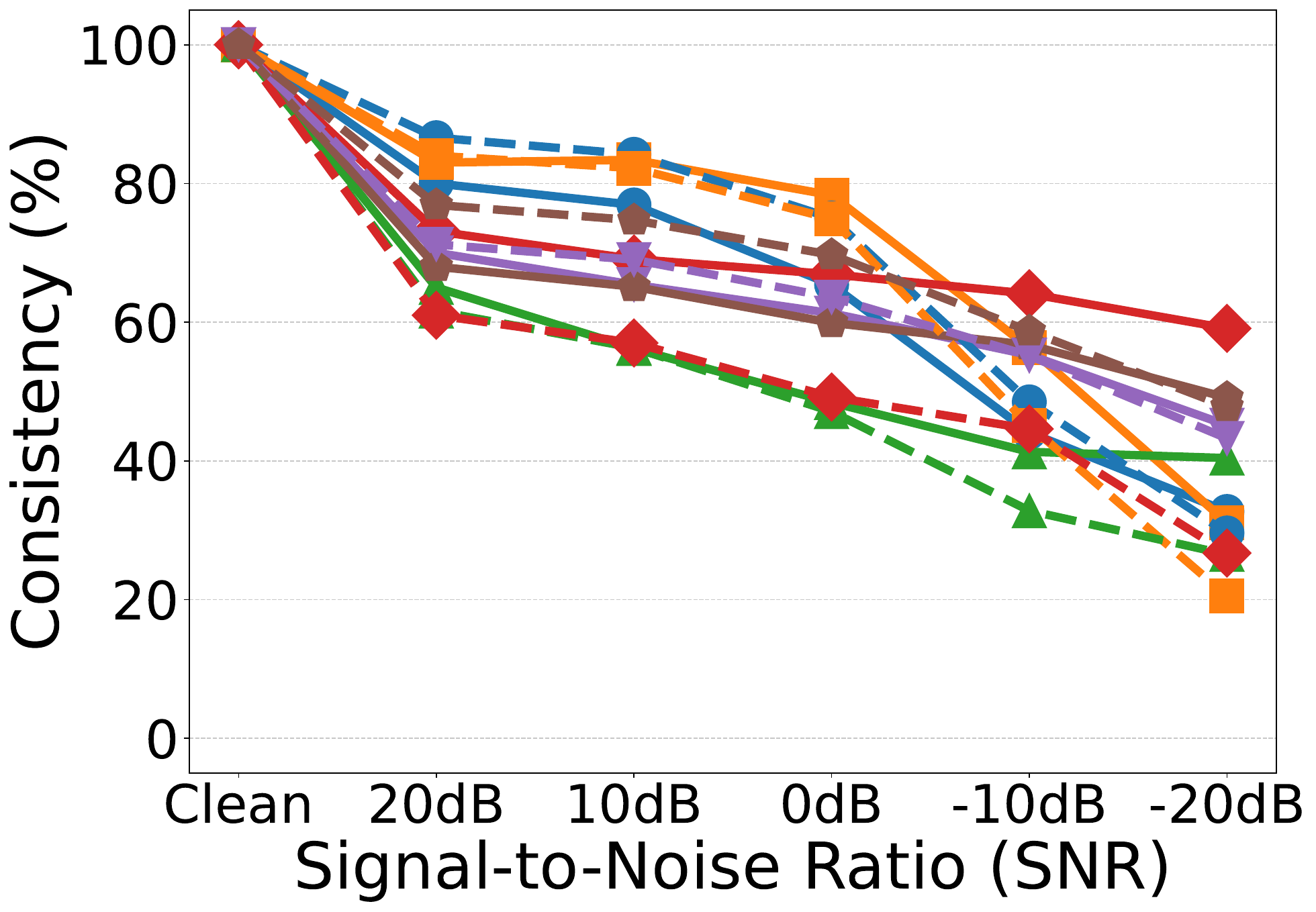}
    \hfill
    \vrule width 0.8pt % Solid vertical line
    \hfill
    \includegraphics[width=\figsizeiii\linewidth]{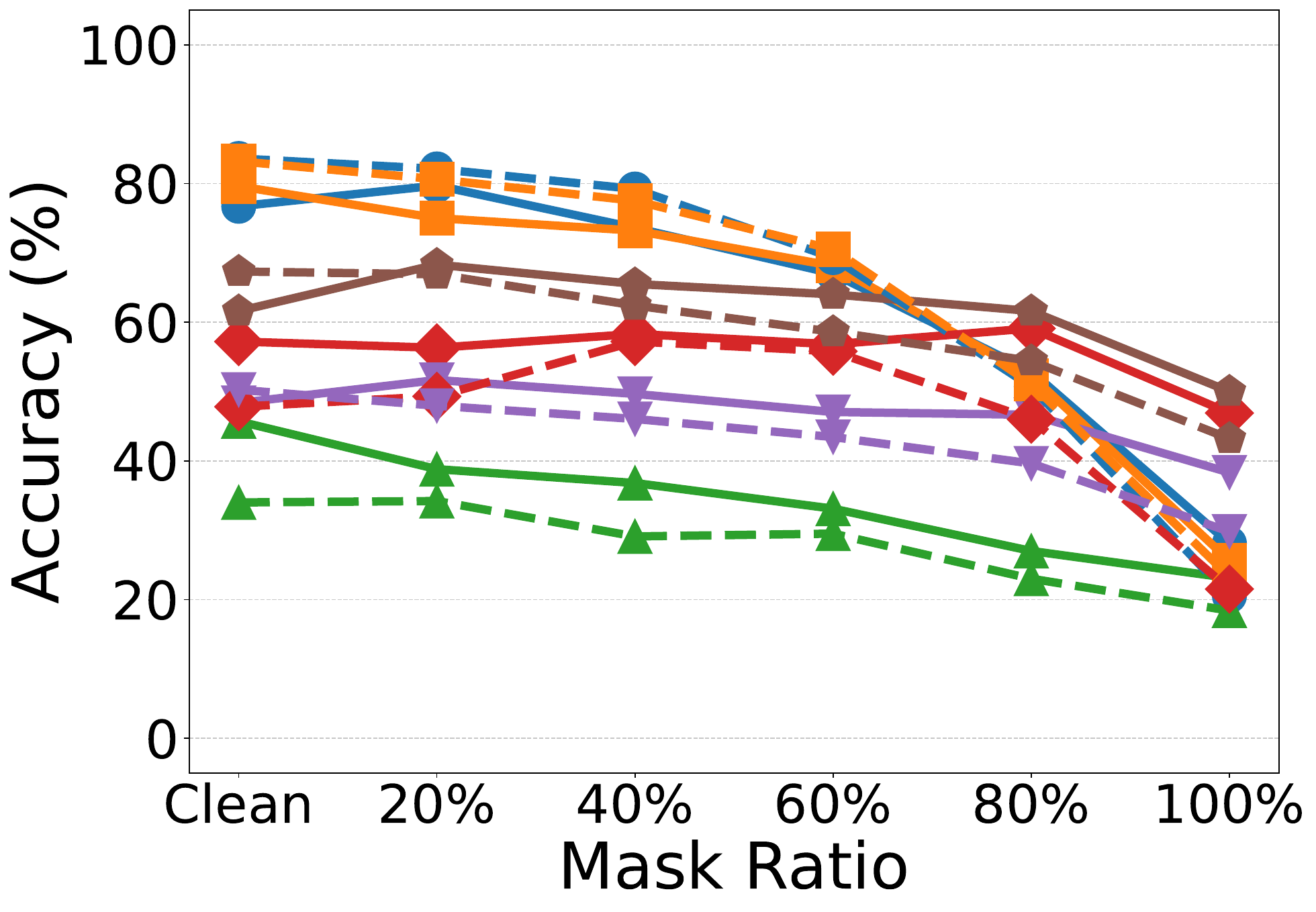}
    \hfill
    \includegraphics[width=\figsizeiii\linewidth]{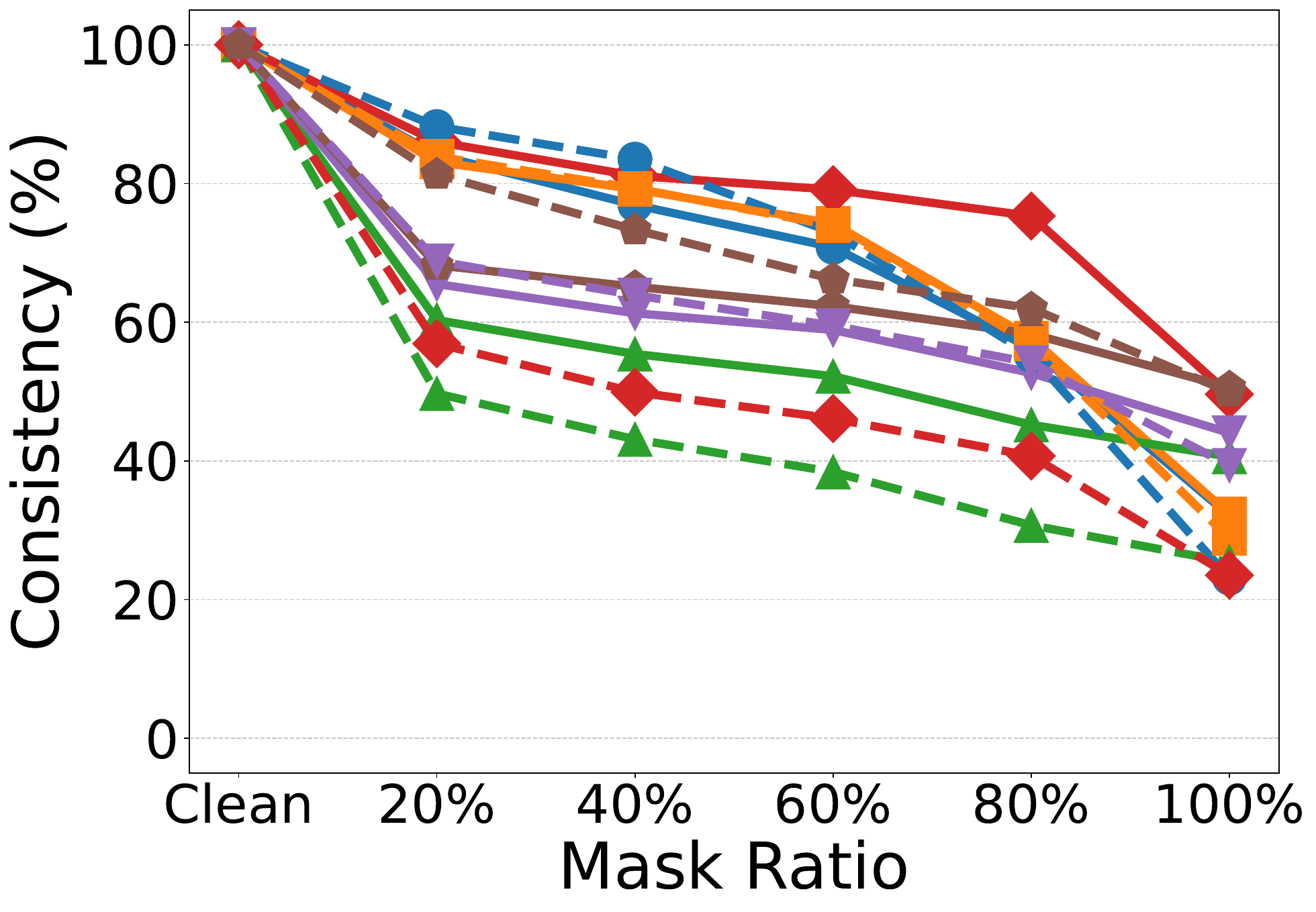}

    \caption{\textbf{Quantitative Impact of Audio Interventions.} The \textbf{(left)} two plots show Noise performance across SNR levels, while the \textbf{(right)} two plots show Masking trends across ratios. Solid lines represent \textit{AF3-Think} and dashed lines represent \textit{Qwen2.5-Omni}.}
    \vspace{-.4cm}
    \label{fig:full_results_quantitative audio}
\end{figure*}

In machine learning, \textit{faithfulness} refers to whether an explanation reflects the model’s actual reasoning process. A faithful explanation correctly shows why the model produced a specific answer. An unfaithful explanation may sound plausible, but it does not match the true decision process. Faithfulness is therefore crucial for building reliable and safe AI systems. Recent work suggests that for text-only LLMs, CoT representations may not reflect the model’s underlying reasoning~\cite{turpin2023languagemodelsdontsay,barez-chain-2025,arcuschin2025chainofthoughtreasoningwildfaithful,chen2023modelsexplainthemselvescounterfactual}. Other studies have proposed methods to measure the faithfulness of CoT explanations~\cite{lanham2023measuringfaithfulnesschainofthoughtreasoning,arcuschin2025chainofthoughtreasoningwildfaithful,matton2025walktalkmeasuringfaithfulness,huang2023largelanguagemodelsexplain,madsen2024selfexplanationslargelanguagemodels}. In the audio domain, several LALMs incorporate chain-of-thought (CoT) reasoning to improve perception and reasoning over audio~\cite{yang2025sakuramultihopreasoninglarge,goel2025audioflamingo3advancing,xie2025audioreasonerimprovingreasoningcapability,chu2024qwen2audiotechnicalreport,kong2025audioflamingosoundcottechnical,ma2025audio}. Although these works report accuracy gains from reasoning, it is unclear whether such reasoning can serve as faithful explanations of the model’s decision process. This question is important because reasoning in audio-language models is inherently more challenging than in text-only models. Despite recent progress, even the most advanced LALMs underperform on expert-level reasoning tasks compared to foundational tasks such as event classification~\cite{yang2025sakuramultihopreasoninglarge}.

%While LLMs have shown strong reasoning abilities through language, extending these systems to understand audio is essential for building models that can reason using contextual auditory cues. 

For text-only LLMs \cite{lanham2023measuringfaithfulnesschainofthoughtreasoning} posits three reasons why CoT may fail as a faithful explanation:  
(i) \textbf{Post-hoc reasoning}: The model may generate reasoning after it has already decided on an answer~\cite{holzinger2017needbuildexplainableai}. Since this reasoning does not influence the decision, it may not reflect the true internal process.  
(ii) \textbf{Extra test-time computation}: The performance gain may come from the extra computation allowed by generating more tokens between the question and the answer~\cite{wei2023chainofthoughtpromptingelicitsreasoning}.  
(iii) \textbf{Encoded reasoning in CoT}: The model may encode useful information (steganography) in ways not understandable to humans, that would encode the answer. This could involve subtle changes in wording, punctuation, or phrasing.

Because Large Audio-Language Models (LALMs) operate conditional to the input sound, it remains unclear whether empirical findings on the faithfulness of CoT representations in text-only LLMs extend to LALMs. In particular due to the additional audio conditioning of LALMs, it is unclear whether the audio makes a genuine impact on the produced CoT, or it in fact consists of hallucinations (i - \textbf{Hallucination-free listening}). If the input audio does have an influence, it is still unclear whether the model exhibits an attention sink type \cite{wang2025mirageeyeshallucinationattack, xiao2024efficientstreaminglanguagemodels, huang2024operaalleviatinghallucinationmultimodal} behavior on the audio, focusing only on a narrow portion, rather than holistically listening to the input (ii - \textbf{holistic listening}). It is also unclear whether the model robustly listens to the input audio following the directives in the question, or whether LALMs prioritize transcriptions over acoustic clues~\cite{chen2025audio} (iii - \textbf{Attentive listening}).

In this paper, we present a faithfulness analysis for CoTs with respect to the audio and also with respect to the output. We provide our analysis on two popular and powerful LALMs, including Qwen2.5 Omni \cite{Qwen2.5-Omni} and AudioFlamingo 3 \cite{goel2025audioflamingo3advancing}. We propose three different audio interventions in order to measure the three qualities of listening we defined above, and we also for the first time incorporate the previously defined CoT interventions on LALMs. 

\section{Research Questions}

% Our goal in this paper is to establish the faithfulness of CoTs produced by LALMs, with respect to i) Input Audio, ii) Final Answer.  As established in the introduction for input audio faithfulness we identify three qualities that are necessary for having a faithful CoT with respect to input audio.

 The goal of this work is to assess the faithfulness of CoT reasoning in LALMs with respect to (i) the input audio and (ii) the final answer. As discussed earlier, we identify three key properties that characterize faithful CoT reasoning with respect to  the audio input.

%We evaluate the faithfulness of LALMs by measuring the consistency of their outputs under various acoustic perturbations. Specifically, we compare the final predictions and consistency in CoT reasoning generated from the intervened audio against a baseline of the original (non-intervened) audio. To evaluate LALM faithfulness, we formulate the following questions:

%\begin{itemize}
%    \item \textbf{RQ1: Robustness to Noise.} How resilient are LALMs to environmental interference? We apply the \textit{Noise Intervention} and compare the resulting CoT and answer consistency against the baseline. 
%    \item \textbf{RQ2: Information Localization.} Do LALMs rely on specific segments or the full audio context? We use \textit{Random Masking} for general testing and \textit{Guided Masking} on a challenging co-reasoning dataset that requires global audio context to correctly answer
%    \item \textbf{RQ3: Acoustic vs. Linguistic Reliance.} Do LALMs prioritize transcriptions over acoustic clues? We generate an \textit{Adversarial Dataset} to test if the CoT remains grounded in the audio when presented with conflicting linguistic cues.
%\end{itemize}

\begin{itemize}
    \item \textbf{Q1: Hallucination-free listening} We ask the question whether the LALM infact incorporates the audio into its answer. To answer this question, we investigate the behavior at the extremes. When contaminated with noise, or given silence, does the same prompt make the model produce an hallucinatory CoT?  
    
    %How resilient are LALMs to environmental interference? We apply the \textit{Noise Intervention} and compare the resulting CoT and answer consistency against the baseline. 
    
    \item \textbf{Q2: Holistic listening} Does the LALM rely on specific segments or the full audio context? Is there an attention sink behavior in the way LALM listens? We use \textit{Random Masking} for general testing and \textit{Guided Masking} on a challenging co-reasoning dataset \cite{wang2025they} that requires global audio context to correctly answer.
    
    \item \textbf{Q3: Attentive Listening} Does the LALM follow the instructions, and listen to what it is supposed to? In order to test for this, we inject an adversarial speech signal into the input audio, mentioning the answer, and see if this adversarial injection in fact changes the answer or not.  
    
    %do the LALM prioritize transcriptions over acoustic clues? We generate an \textit{Adversarial Dataset} to test if the CoT remains grounded in the audio when presented with conflicting linguistic cues.
    
    \item \textbf{Q4: CoT-Output Faithfulness} We finally test whether the CoTs that the LALM produce are in fact faithful to the produced model output. For this purpose we apply the interventions introduced in \cite{lanham2023measuringfaithfulnesschainofthoughtreasoning}, which check for posthoc reasoning, extra test-time computation, and encoded reasoning, as mentioned in the introduction. 
\end{itemize}
%We describe the intervention pipeline in Figure \ref{fig:full_pipeline_centered}. We elaborate more en each intervention, and the obtained results in subsequent sections as well. 

\vspace{-.3cm}
\section{Audio Interventions}
To avoid the unintended out-of-distribution (OOD) artifacts commonly introduced by audio editing tools, we developed "context-preserving" interventions. Next, we detail the experimental setup used to apply these audio modifications. \\
\noindent \textbf{Benchmarks:} We perform our audio interventions evaluate on three diverse datasets:
\begin{itemize}
    \item \textbf{SAKURA:} \cite{yang2025sakuramultihopreasoninglarge} Tests single- and multi-hop reasoning over 500 multiple-choice questions per track, focusing on gender, language, emotion, and animal sounds.
    \item \textbf{MMAR:} \cite{ma2025mmarchallengingbenchmarkdeep} A challenging benchmark with 1,000 triplets from real-world videos. It requires multi-step reasoning across mixtures of speech, music, and environmental audio.
    \item \textbf{MMAU:} \cite{sakshi2024mmaumassivemultitaskaudio} Evaluates expert-level multimodal understanding using 1,000 curated clips paired with human-annotated questions and answers.

\end{itemize}

\begin{figure*}[t]
    \centering
    % Legend Row
    \includegraphics[width=0.7\linewidth]{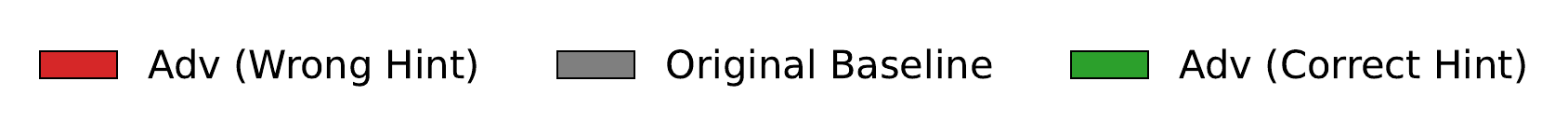}
    
    \vspace{-.2cm} 

    % AF3 Group
    \begin{subfigure}{0.43\textwidth}
        \centering
        \textbf{Audio Flamingo 3 (AF3)}\\[0.5em]
        \includegraphics[width=0.45\linewidth]{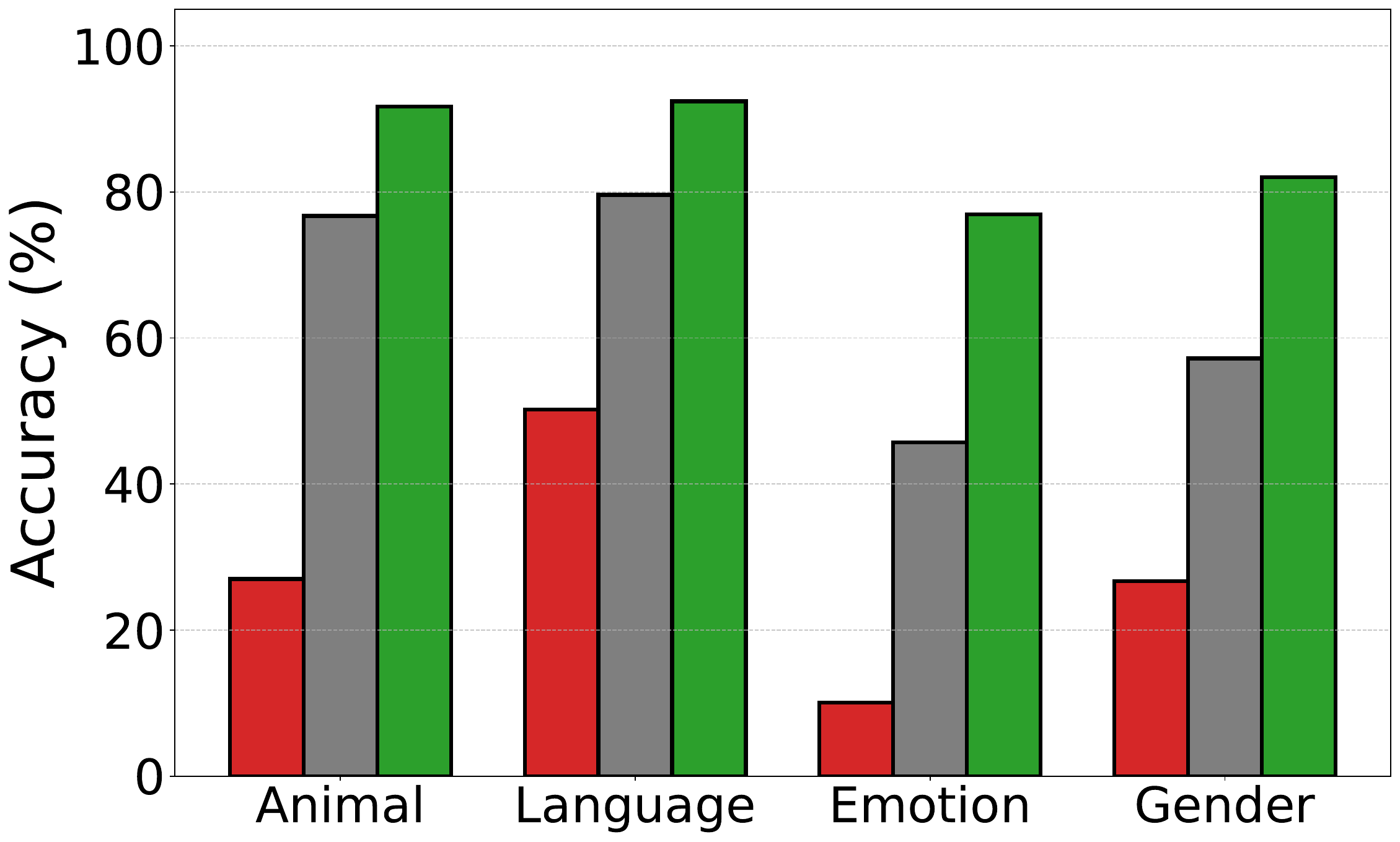}
        \hfill
        \includegraphics[width=0.45\linewidth]{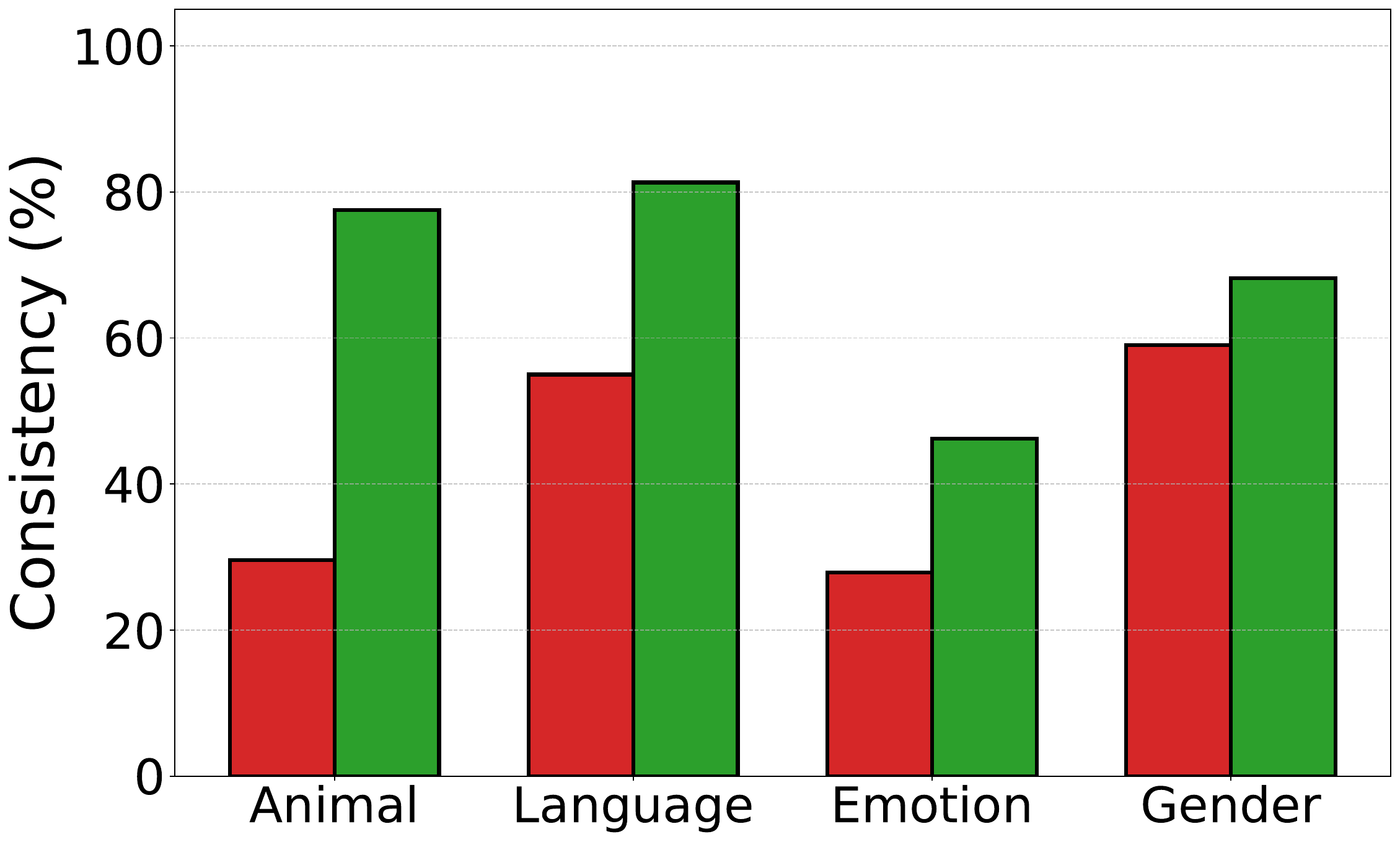}
    \end{subfigure}
    \hfill
    \vrule width 0.8pt % Solid vertical separator
    \hfill
    % Qwen Group
    \begin{subfigure}{0.43\textwidth}
        \centering
        \textbf{Qwen2.5-Omni}\\[0.5em]
        \includegraphics[width=0.45\linewidth]{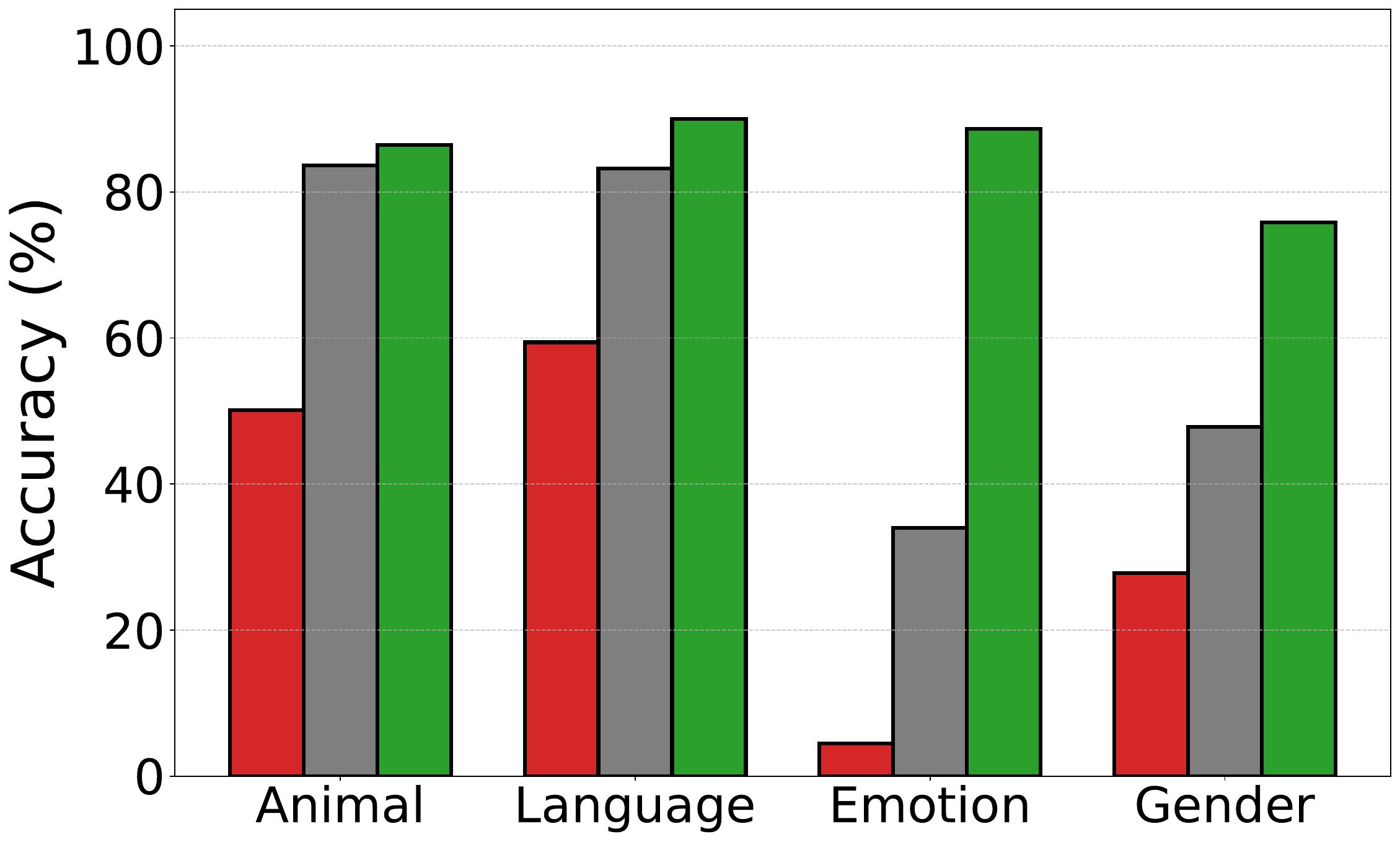}
        \hfill
        \includegraphics[width=0.45\linewidth]{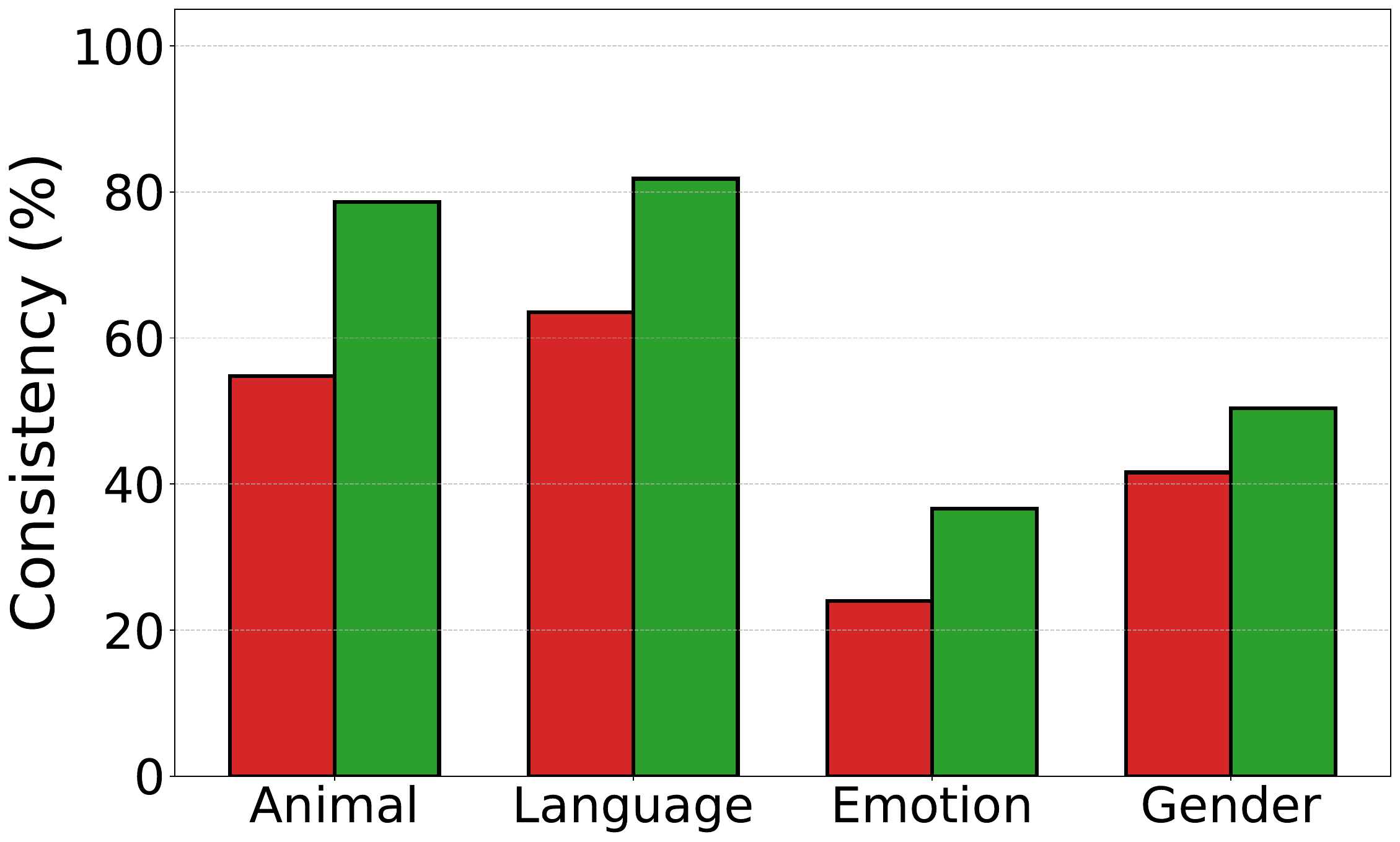}
    \end{subfigure}
    \vspace{-.3cm}
    \caption{\textbf{Adversarial Intervention Results (RQ3).} Accuracy and CoT Consistency for \textit{AF 3} \textbf{(left)} and \textit{Qwen2.5-Omni} \textbf{(right)}. }
    \vspace{-.7cm}
    \label{fig:adv_results}
\end{figure*}

\noindent \textbf{Models:} We utilize two high-performance open-source models capable of structured reasoning: \textit{Audio Flamingo 3-Think} \cite{goel2025audioflamingo3advancing} and \textit{Qwen2.5-Omni} \cite{Qwen2.5-Omni}.

\noindent \textbf{Evaluation Procedure:} As shown in Figure \ref{fig:pipeline_cot_intervention}, we prompt the models to "think and reason step-by-step" before providing a final answer. We use automated preprocessing to isolate the reasoning string from the final prediction. Consistency is evaluated in two parts:
\begin{enumerate}
    \item \textbf{Answer Consistency:} We compare the final prediction of the noisy audio against the clean baseline answer to calculate accuracy trends.
    \item \textbf{CoT Consistency:} We employ an \textit{LLM-as-a-judge} (\texttt{Mistral-Small-3.1-24B-Instruct-2503}) to rate the semantic similarity between the baseline and intervened reasoning on a 1--5 scale. A score of 5 represents \textit{Perfectly Consistent} (identical meaning), while a score of 1 represents \textit{Contradictory} (opposing conclusions).
\end{enumerate}
We elaborate more on each intervention, and the obtained results in the following subsections.

\subsection{Adding Noise}
\label{sec:study1}

This intervention enables assessing the Q1-Hallucination-free listening, as well as robustness of the model to noise (Q2-Attentive Listening).  

\noindent \textbf{Setup:} We overlay Gaussian white noise on the entire audio signal at five Signal-to-Noise Ratio (SNR) levels: $\{-20, -10, 0, 10, 20\}$ dB. At the extreme $-20$ dB level, the audio is nearly indistinguishable from pure noise. We anticipate that models will either hallucinate or fail to provide a valid response at this threshold.

%This study evaluates the resilience of LALMs to environmental acoustic interference by measuring the consistency of reasoning and predictions under varying noise levels.
%\subsection{Experimental Setup}

\noindent \textbf{Results:} In the first and second panels of Figure \ref{fig:full_results_quantitative audio} we illustrate accuracy and consistency trends (respectively) for final predictions and reasoning chains across different SNR values. We observe that both AF3 and QWEN2.5 retain their performance until 0dB SNR, after which the drops in performance and consistency become severe. 

In Table \ref{tab:cot_consistency}, we observe that in fact depending on the dataset the models exhibit hallucinatory CoT, which talks about non-existent phenomena in audio. For instance on MMAU, AF3 obtains a CoT consistency score of 3.57, even for -20dB SNR noise contaminated input audio. In contrast, for instance, QWEN gives reasoning chains that indicate - it cannot hear any audible signal in the input audio, which results in low consistency scores (E.g. QWEN on SAKURA animal).

%\begin{table}[t]
%\centering
%\tiny
%\begin{tabular}{l |l | c c c c c c}
%\hline
%\textbf{Intervention} & \textbf{Model} & \textbf{Animal} & \textbf{Language} & \textbf{Gender} & \textbf{Emotion} & \textbf{MMAR} & \textbf{MMAU} \\
%\hline
%
%Masking 100\% & AF3  & 3.01 & 2.63 & 2.87 & 3.10 & 3.19 & 3.65 \\
%             & Qwen & 2.35   & 2.40   & 2.95   & 2.64   & 2.8129 & 3.39   \\
%             \midrule
%
%Masking 20\%  & AF3  & 4.60 & 4.32 & 4.353 & 4.043 & 3.97 & 4.41 \\
%             & Qwen & 4.68   & 4.62   & 3.89   & 3.45   & 4.01   & 4.45   \\
%\midrule
%
%Noise -20dB      & AF3  & 2.89 & 2.54 & 3.17 & 3.13 & 3.09 & 3.57 \\
%             & Qwen & 2.45   & 2.27   & 2.70   & 2.61   & 2.82   & 3.22 \\
%\midrule
%SNR +20      & AF3  & 4.41 & 4.27 & 3.76 & 4.22 & 4.12 & 4.44 \\
%             & Qwen & 4.58   & 4.67   & 3.93   & 4.05   & 4.12   & 4.33 \\
%             \midrule
%Adv-correct  & AF3  & 4.25 & 4.26 & 3.59 & 3.55 & -- & -- \\
%             & Qwen & 4.46   & 4.59   & 3.23   & 2.88   & -- & -- \\
%\midrule
%Adv-wrong    & AF3  & 3.04 & 3.43 & 3.32 & 2.99 & -- & -- \\
%             & Qwen & 3.56   & 3.86   & 2.92   & 2.49   & -- & -- \\
%
%\hline
%\end{tabular}
%\caption{The CoT intervention consistencies}
%\label{tab:cot_consistency}
%\end{table}
%

\begin{table}[htb]
\centering
\resizebox{.9\linewidth}{!}{
\large% Use small font size
\setlength{\tabcolsep}{5pt} % Reduce column spacing
\renewcommand{\arraystretch}{1.3} % Adjust row height
\begin{tabular}{l | l | c  c  c  c  c  c}
%\hline
\textbf{Intervention} & \textbf{Model} & \textbf{Animal} & \textbf{Language} & \textbf{Gender} & \textbf{Emotion} & \textbf{MMAR} & \textbf{MMAU} \\
\hline \hline
\multirow{2}{*}{Mask 100\%} & AF3  & 3.01 & 2.63 & 2.87 & 3.10 & 3.19 & 3.65\\
                              & Qwen & 2.35 & 2.40 & 2.95 & 2.64 & 2.81 & 3.39\\
 \hline
\multirow{2}{*}{Mask 20\%}  & AF3  & 4.60 & 4.32 & 4.35 & 4.04 & 3.97 & 4.41\\
                              & Qwen & 4.68 & 4.62 & 3.89 & 3.45 & 4.01 & 4.45\\
\hline
\multirow{2}{*}{-20dB SNR}      & AF3  & 2.89 & 2.54 & 3.17 & 3.13 & 3.09 & 3.57\\
                              & Qwen & 2.45 & 2.27 & 2.70 & 2.61 & 2.82 & 3.22\\
\hline
\multirow{2}{*}{20dB SNR}      & AF3  & 4.41 & 4.27 & 3.76 & 4.22 & 4.12 & 4.44\\
                              & Qwen & 4.58 & 4.67 & 3.93 & 4.05 & 4.12 & 4.33\\
\hline
\multirow{2}{*}{Adv-correct}  & AF3  & 4.25 & 4.26 & 3.59 & 3.55 & N.A.    & N.A.   \\
                              & Qwen & 4.46 & 4.59 & 3.23 & 2.88 & N.A.    & N.A.   \\
\hline
\multirow{2}{*}{Adv-wrong}    & AF3  & 3.04 & 3.43 & 3.32 & 2.99 & N.A.    & N.A.   \\
                              & Qwen & 3.56 & 3.86 & 2.92 & 2.49 & N.A.    & N.A.   \\
%\hline
\end{tabular}
}
\caption{Audio intervention CoT consistencies}
\vspace{-1cm}
\label{tab:cot_consistency}
\end{table}

\subsection{Masking}
\label{sec:masking}

To address Q1 (Hallucination-free listening) and Q2 (Holistic listening), we apply \textit{Random Masking} and \textit{Guided Masking} to evaluate if models stay grounded in the actual audio signal. We achieve this by removing or isolating specific components of the audio input and observing the impact on the model's CoT and final predictions.

\noindent \textbf{Setup:} 
For \textit{Random Masking}, we mask [20\%, 40\%, 60\%, 80\%, and 100\%] of the audio in distributed chunks. For tasks focusing on global features, such as predicting an animal, gender, or emotion rather than specific speech content, randomized masking should not disproportionately affect performance unless the model relies on specific "attention sinks" or hidden encoded representations within a narrow segment of the audio. By distributing masks across the entire duration, we ensure the model cannot rely on a single localized "shortcut" and instead must attentively listen to the entire audio context. At 100\% masking (total silence), we evaluate Q1 to determine if the model produces hallucinatory reasoning when no signal is present. 

We further utilize \textit{Guided Masking} on the JASCO dataset \cite{wang2025they} to investigate how models attend to different modalities (speech vs. audio). Following the original evaluation pipeline \cite{wang2025they}, we use an LLM-as-a-judge (\texttt{Mistral-Small-3.1-24B-Instruct-2503}) with a three-tier scoring system to compare model predictions against reference answers. To determine modality dependency, the judge evaluates whether the model’s final reasoning is drawn from both modalities or a single source. By specifically masking either the speech segments (\textit{Speech Mask}) or the general background sounds (\textit{Audio Mask}), we force the model to shift its reasoning based on the remaining modality, testing whether it achieves true joint audio-speech understanding or relies on a single dominant source.

\newcommand{\figsize}{0.19}
\begin{figure*}[t]
    \centering
    % Row 0: Legend
    \includegraphics[width=0.8\linewidth]{figures/audio_intervention/legend_datasets_only.pdf}
    
    \vspace{-.3cm}

    % Row 1 Header: Labels for Columns
    %\begin{minipage}{0.24\textwidth}\centering \small \textbf{Paraphrasing}\end{minipage}
    %\begin{minipage}{0.24\textwidth}\centering \small \textbf{Early Answering}\end{minipage}
    %\begin{minipage}{0.24\textwidth}\centering \small \textbf{Adding Mistake}\end{minipage}
    %\begin{minipage}{0.24\textwidth}\centering \small \textbf{Filler Token}\end{minipage}

    \vspace{-.7cm}

    % Row 1: AF3 Plots
    \rotatebox{90}{\makebox[0.18\textwidth]{\small \textbf{AF3}}} %\hfill
    \includegraphics[width=\figsize\textwidth]{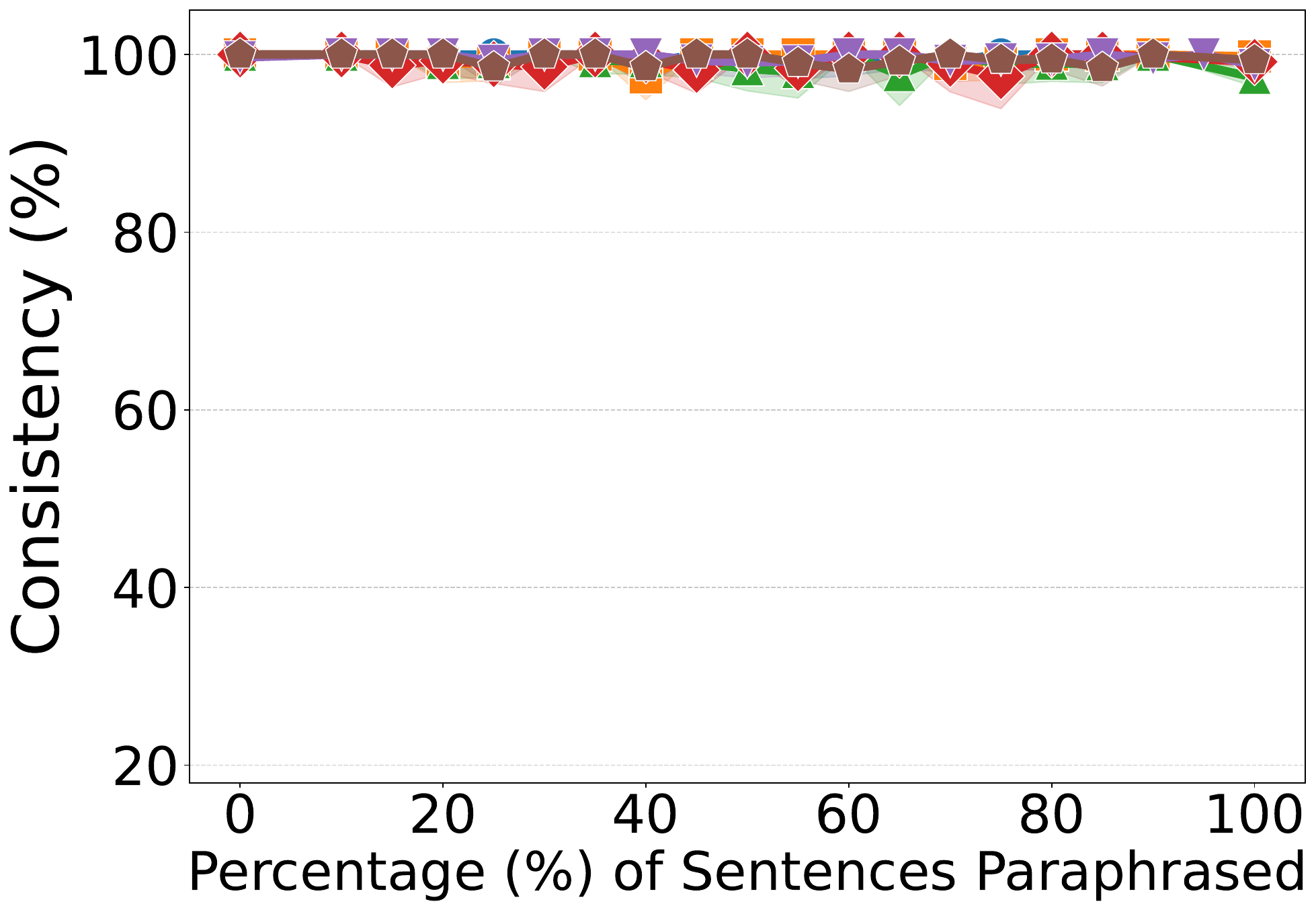} %\hfill
    \includegraphics[width=\figsize\textwidth]{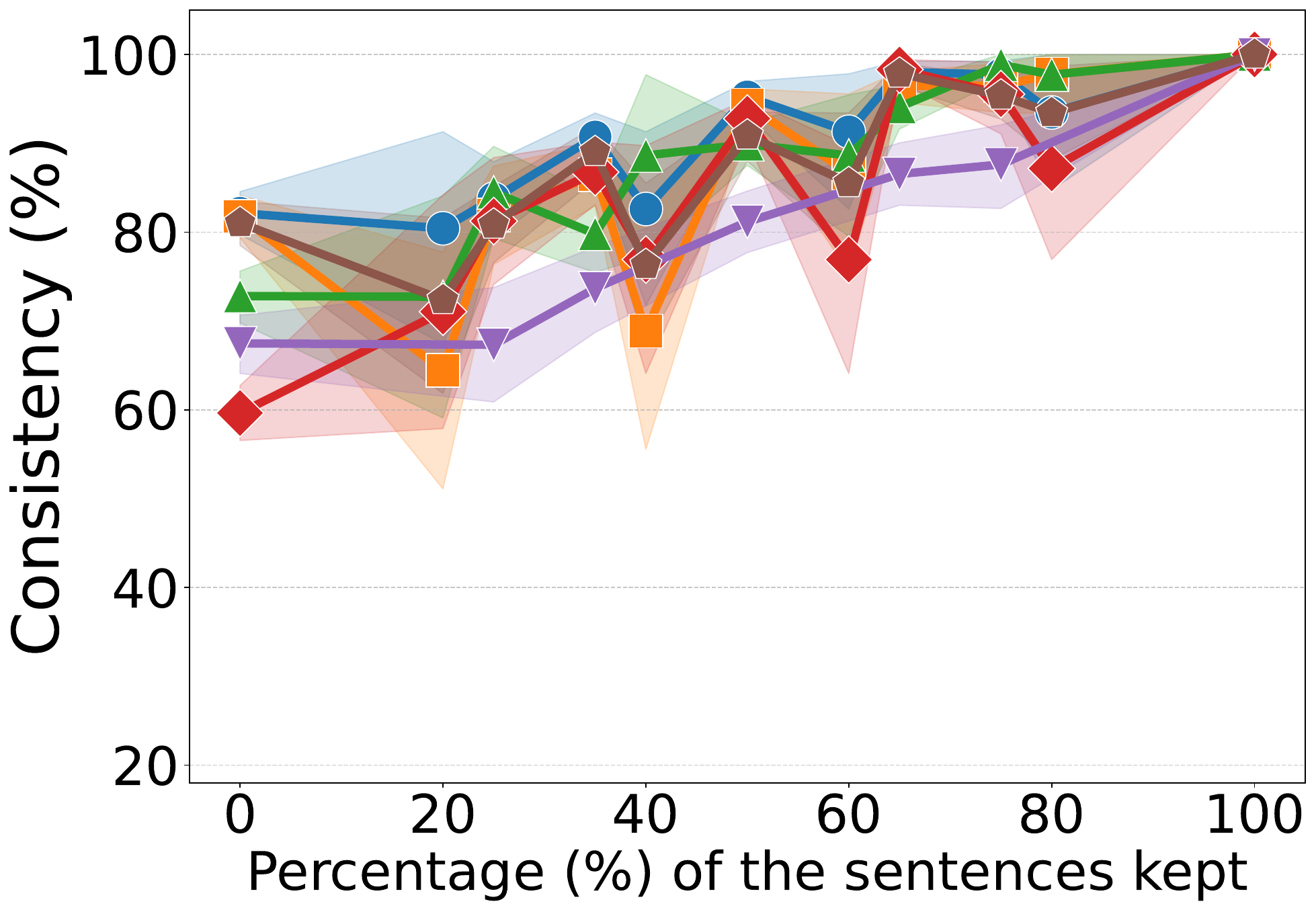} %\hfill
    \includegraphics[width=\figsize\textwidth]{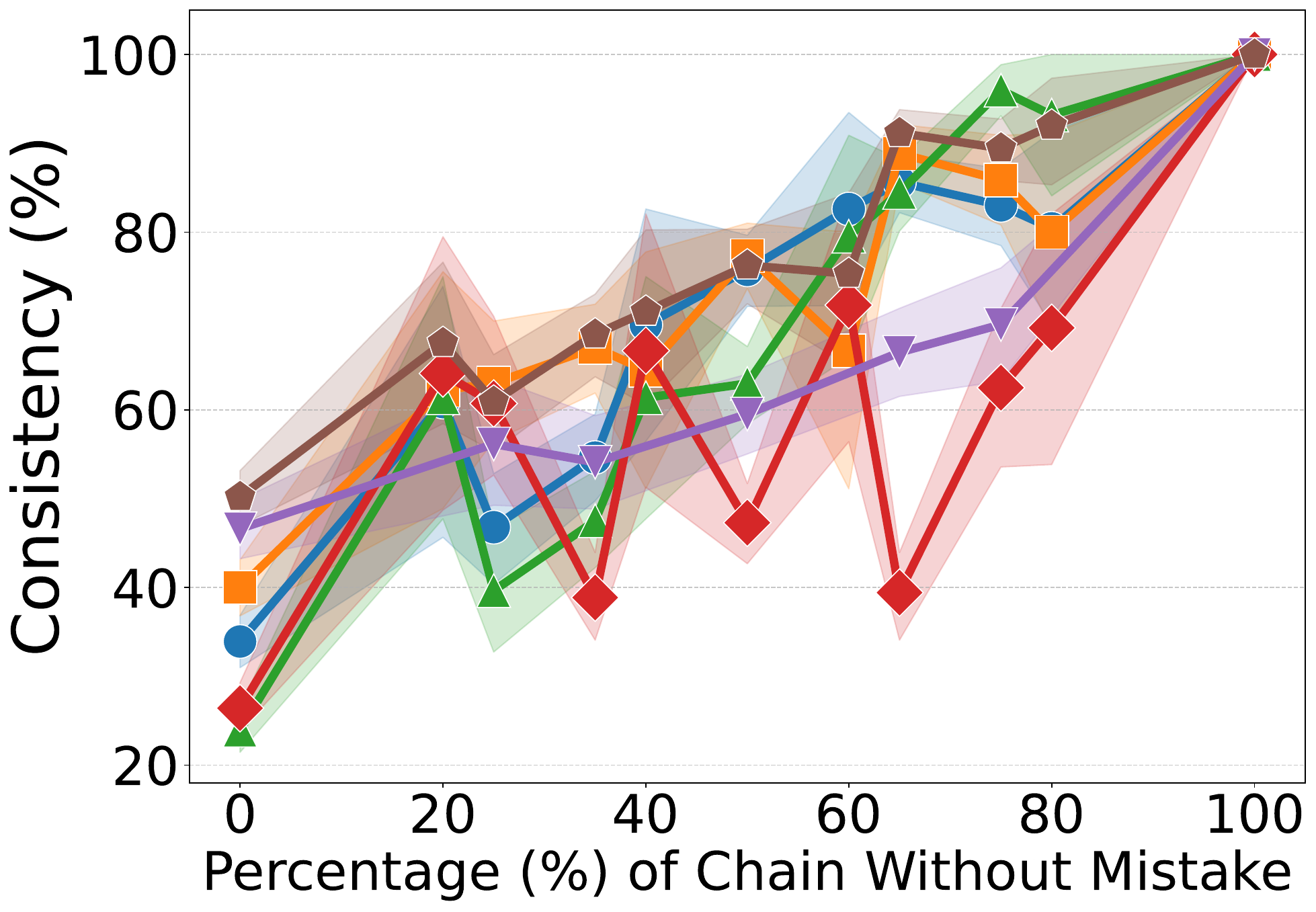} %\hfill
    \includegraphics[width=\figsize\textwidth]{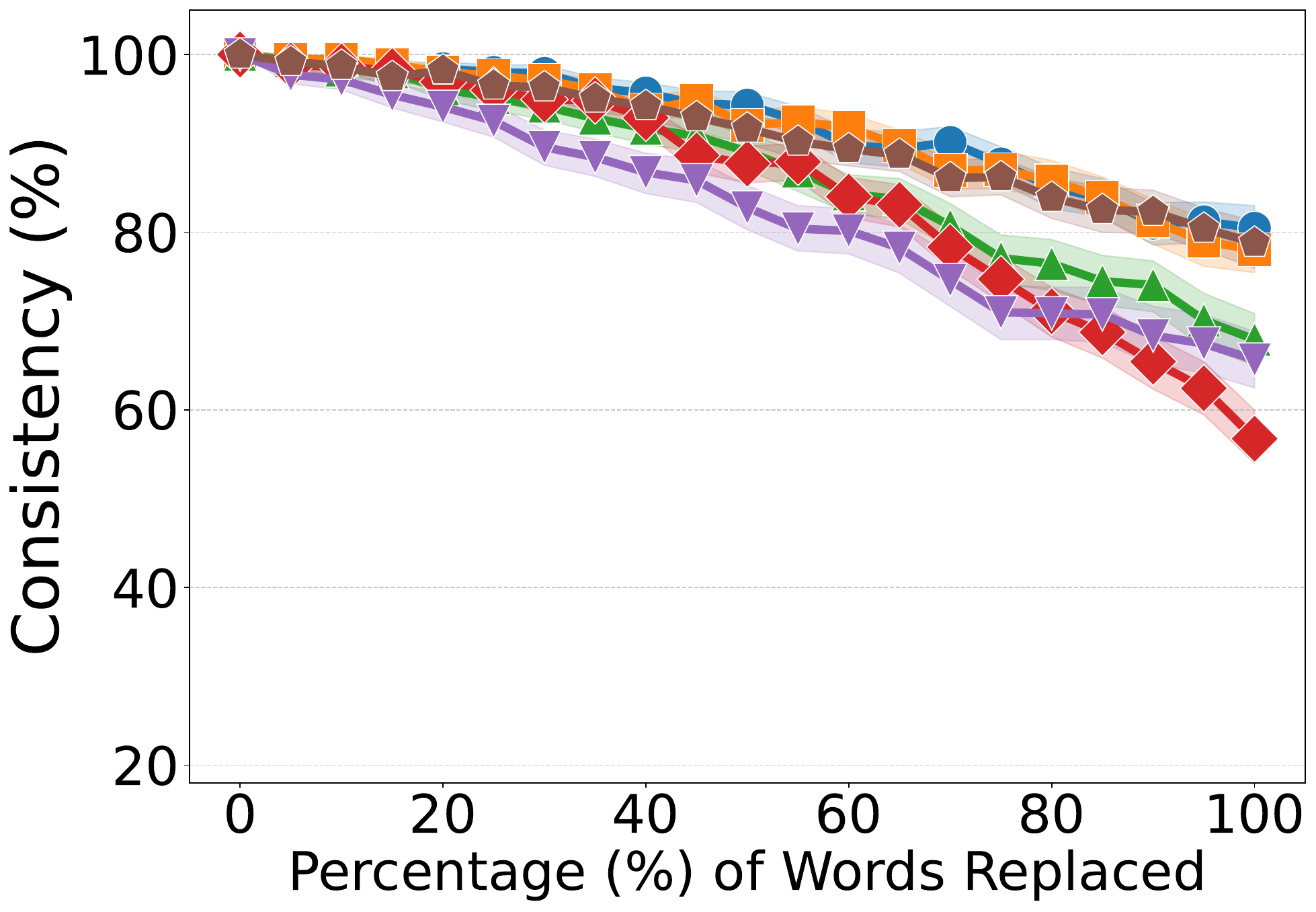}

    \vspace{-0.6cm}

    % Row 2: Qwen2.5-Omni Plots
    \rotatebox{90}{\makebox[0.18\textwidth]{\small \textbf{Qwen2.5}}} %\hfill
    \includegraphics[width=\figsize\textwidth]{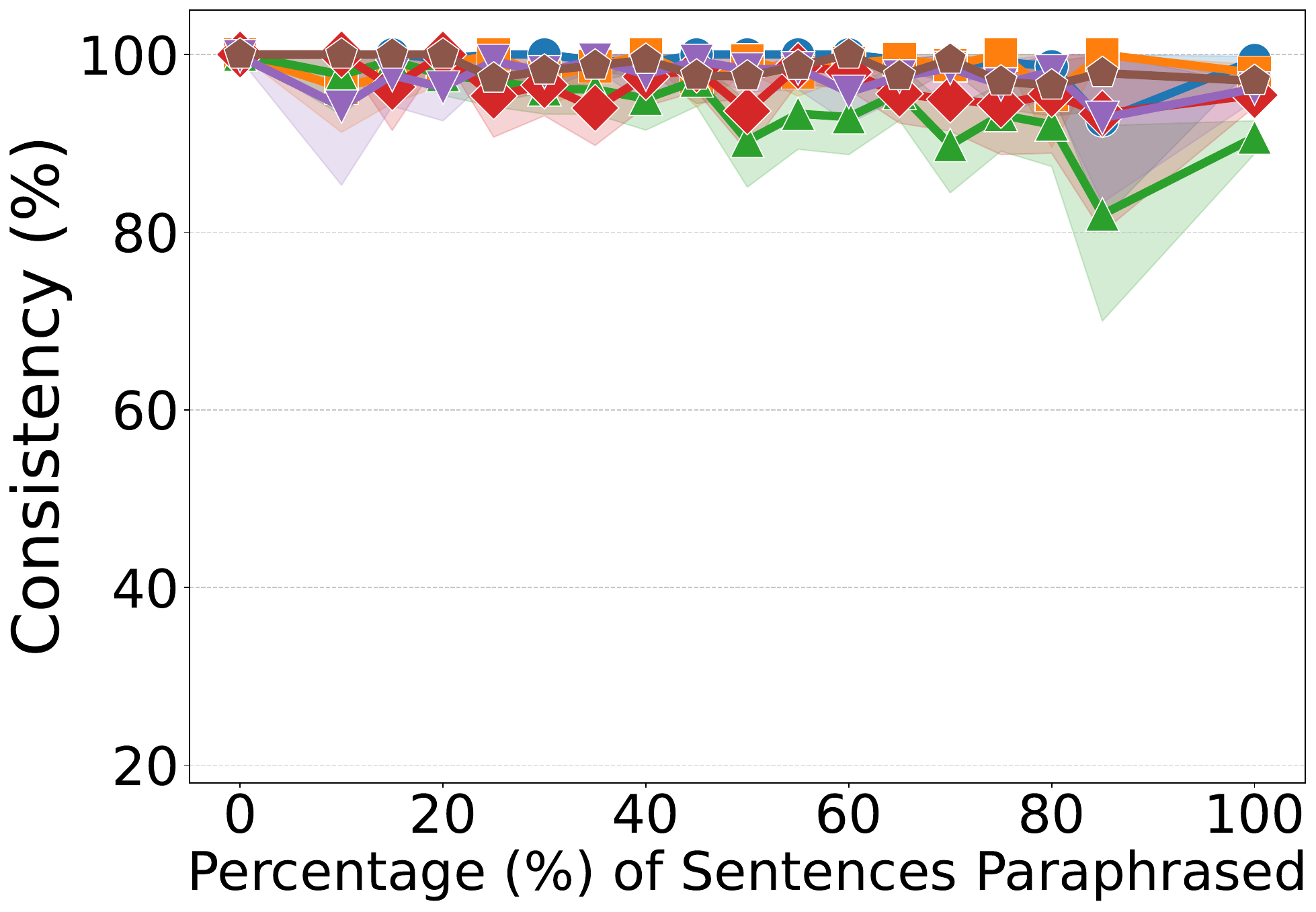} %\hfill
    \includegraphics[width=\figsize\textwidth]{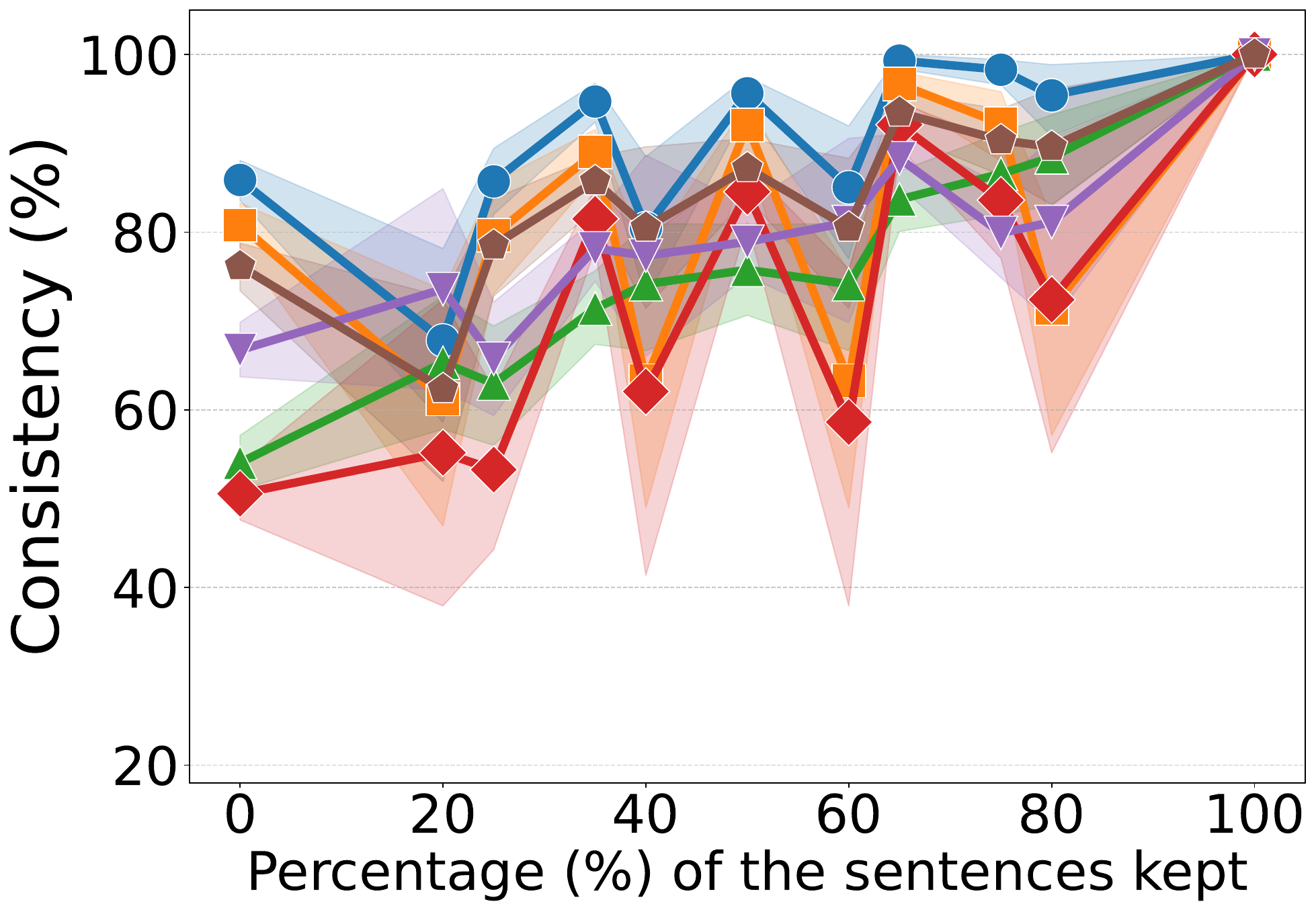} %\hfill
    \includegraphics[width=\figsize\textwidth]{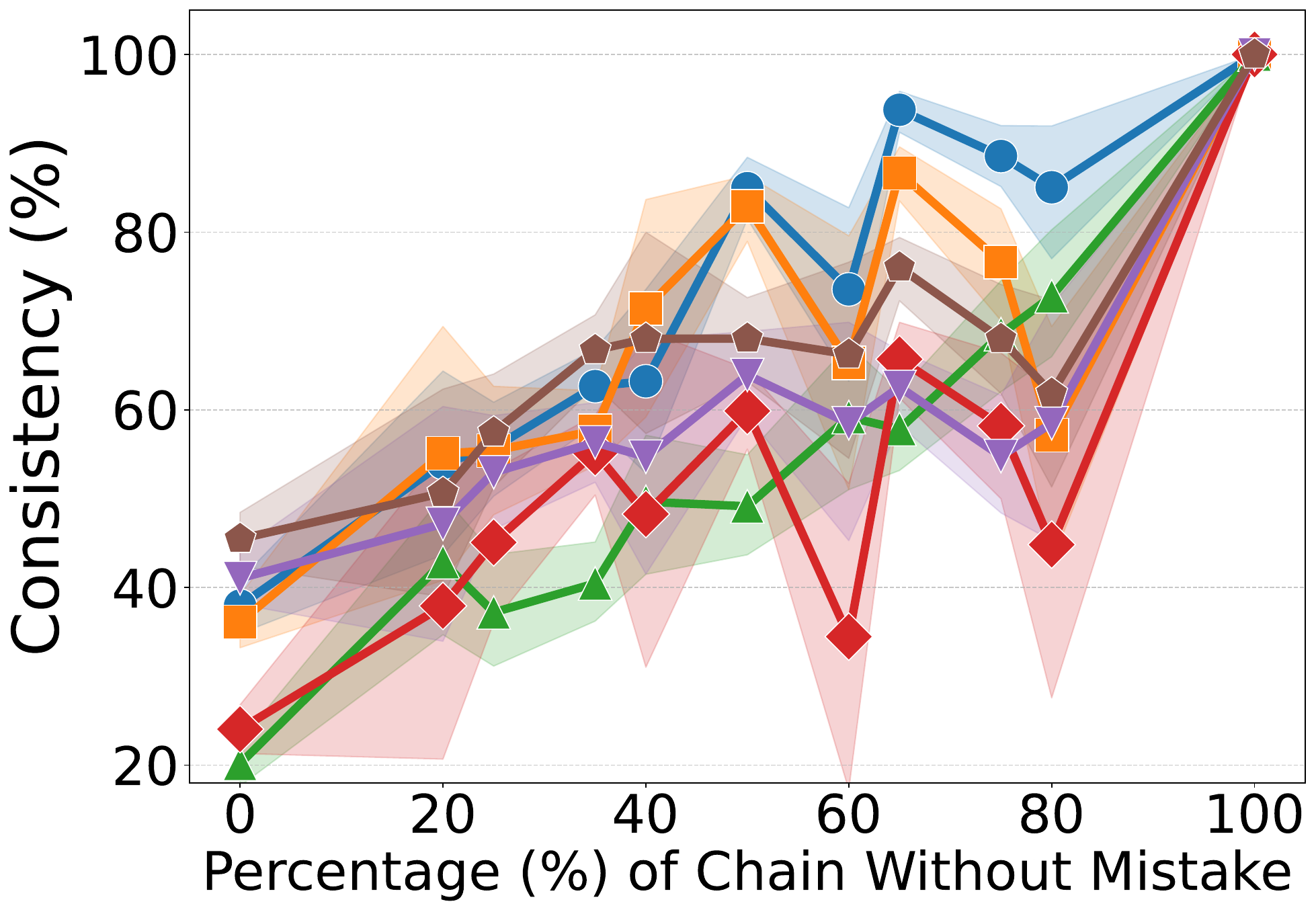}% \hfill
    \includegraphics[width=\figsize\textwidth]{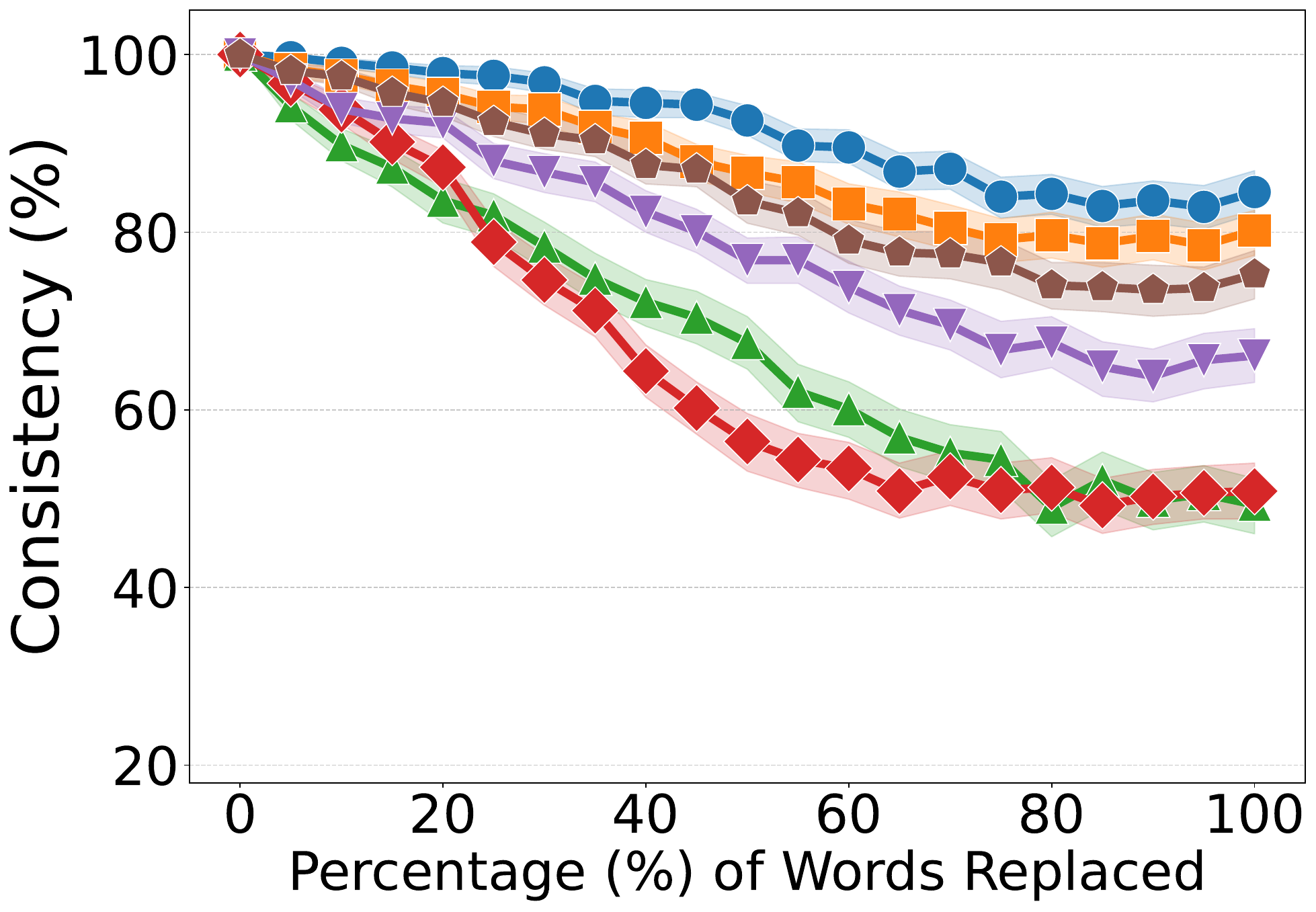}
    \vspace{-.3cm}
    \caption{CoT Interventions. \textbf{(left)} Paraphrasing, \textbf{(mid-left)} Early Answering, \textbf{(mid-right)} Adding Mistakes, \textbf{(right)} Filler Tokens} 
    \vspace{-.4cm}
    \label{fig:cot_intervention_grid}
\end{figure*}

\begin{figure}[htb]
    \centering
    % Left Figure: AF3
    \begin{subfigure}{0.48\columnwidth}
        \centering
        \textbf{Audio Flamingo 3 (AF3)}\\[0.2em]
        \includegraphics[width=\linewidth]{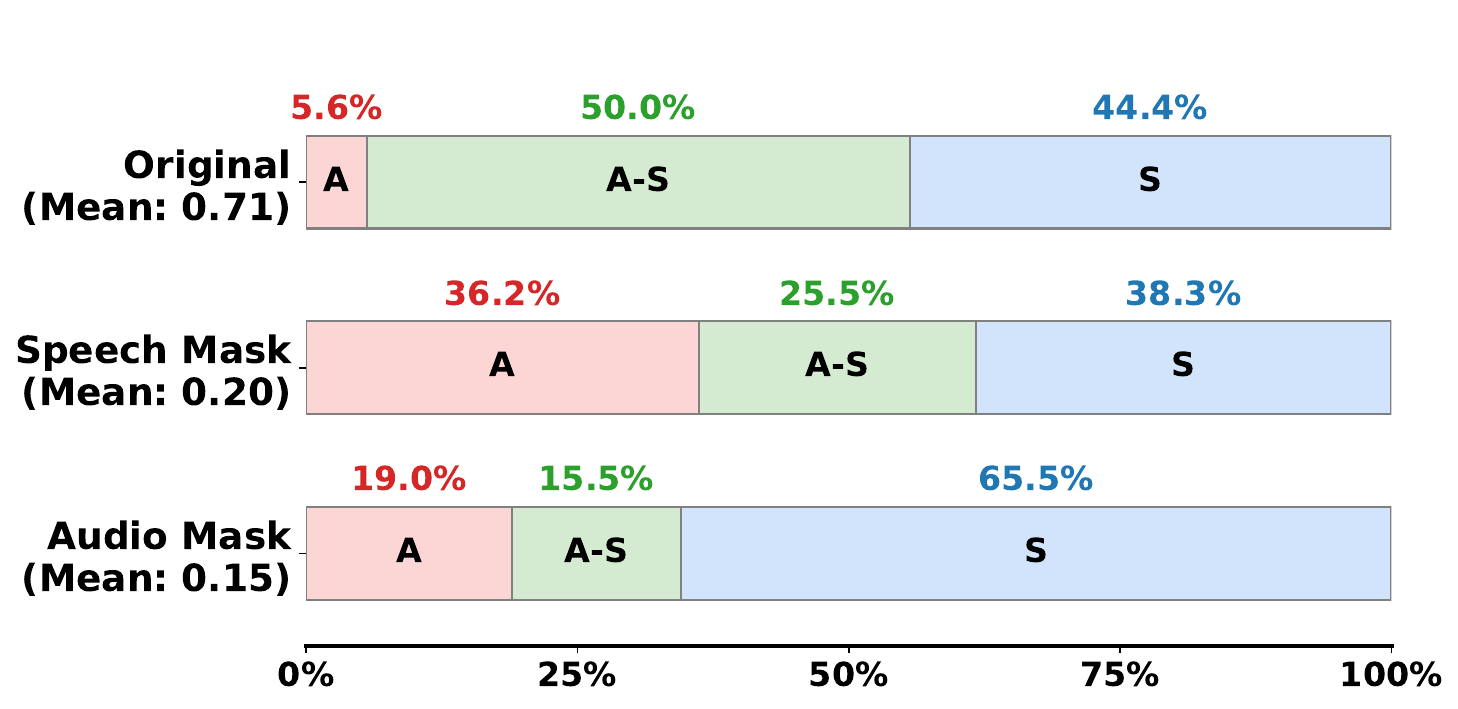}
        \label{fig:af3_bar_scaled}
    \end{subfigure}
    \hfill % This horizontal fill puts them side-by-side
    % Right Figure: Qwen2.5-Omni
    \begin{subfigure}{0.48\columnwidth}
        \centering
        \textbf{Qwen2.5-Omni}\\[0.2em]
        \includegraphics[width=\linewidth]{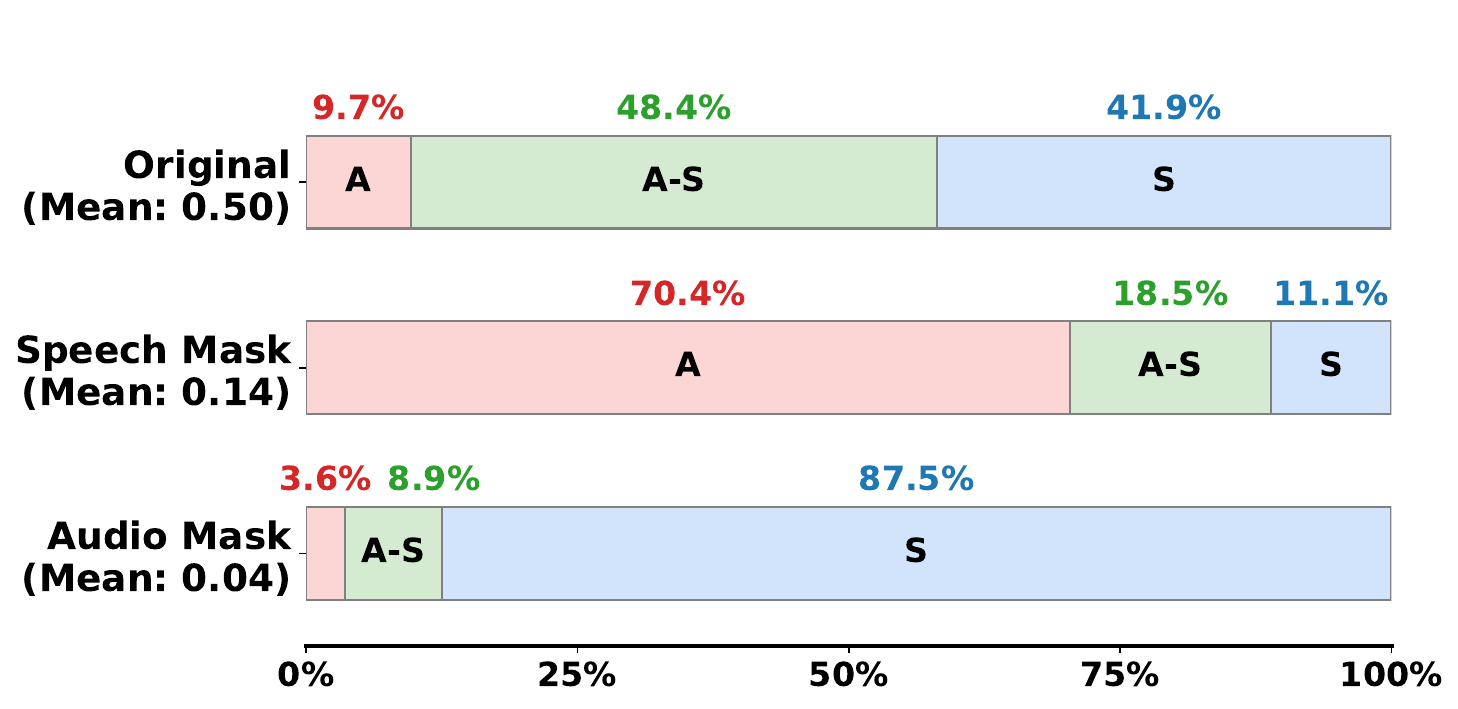}
        \label{fig:qwen_bar_scaled}
    \end{subfigure}
    \vspace{-.7cm}
    \caption{\textbf{Modality Dependency under Guided Masking.} The values in parentheses (on the left) indicate the mean similarity score compared to the reference answers. The percentages above the bars represent the proportion of responses that are audio-dependent (A), both-dependent (A-S), and speech-dependent (S).}
    \vspace{-.6cm}
    \label{fig:jasco_result}
\end{figure}

\noindent \textbf{Results:} 
The accuracy and consistency plots in Figure \ref{fig:full_results_quantitative audio} show that both models remain stable up to a 60\% random masking ratio but fail significantly at higher ratios. Figure \ref{fig:jasco_result} further reveals a clear imbalance in how the models process multimodal inputs: guided masking shows that they do not listen holistically (Q2) and instead rely heavily on speech. The mean similarity scores and modality bars suggest that the models attempt to adjust their reasoning depending on which modality remains available—referring more to audio when speech is masked and vice versa, but performance still drops under any guided intervention. Interestingly, masking general audio while keeping speech leads to lower similarity scores. This likely occurs because the model can more easily hallucinate details from spoken context than infer information from background sounds. Table \ref{tab:cot_consistency} further shows that AF3 often produces hallucinatory reasoning when the input is silent. At 100\% masking, it still maintains a consistency score of 3.65 on MMAU. In contrast, Qwen2.5-Omni remains more grounded and frequently reports that it cannot hear an audible signal. This behavior leads to lower consistency scores (e.g., 2.35 on SAKURA-Animal) but reduces hallucinated reasoning.

    %\caption{\textbf{Modality Dependency under Guided Masking.} The values in parentheses on the left indicate the mean similarity score compared to the reference answers. The percentages above the bars represent the proportion of responses that are audio-dependent (A), both-dependent (A-S), and speech-dependent (S).}
    %\label{fig:jasco_result}
%\end{figure}

% Regarding \textbf{Q1}, Table \ref{tab:cot_consistency} indicates that AF3 frequently hallucinates reasoning for silent inputs. At 100\% masking, AF3 produces a CoT that appears "consistent" with a real answer, scoring 3.65 on MMAU. In contrast, Qwen2.5-Omni is more grounded; its CoT often correctly states it cannot hear an audible signal, leading to lower consistency scores (e.g., 2.35 on SAKURA-Animal).

% Your results text or table goes here.

%\noindent \textbf{Results:} 
%The accuracy and consistency plots in Figure \ref{fig:full_results_quantitative audio} show that while models are stable up to 60\% random masking, they exhibit failures at higher ratios. As shown in Figure \ref{fig:jasco_result}, masking different parts of the audio reveals a clear imbalance in how models listen. For Q2, both models rely much more on speech than on general audio. When we mask only the general audio (keeping speech), Qwen2.5 still yields 87.5\% correct reasoning based solely on speech, while AF3 relies on speech for 65.5\% of its correct outputs. When speech is masked, performance drops significantly for both models (Mean 0.20 for AF3 and 0.14 for Qwen), proving they do not listen "holistically" and are heavily biased toward speech.
%
%Regarding Q1, Table \ref{tab:cot_consistency} shows that AF3 often hallucinates reasoning for silent inputs. At 100\% masking (total silence), AF3 still produces a CoT that appears "consistent" with a real answer, scoring 3.65 on MMAU. In contrast, Qwen2.5 is more honest; its CoT often states it cannot hear anything, leading to lower consistency scores (e.g., 2.35 on SAKURA-Animal) but fewer hallucinations.

\vspace{-.3cm}
\subsection{Adversarial Injection}
\label{sec:rq3}
This intervention assesses for attentive listening (Q3).

\noindent \textbf{Setup:} We generate an \textit{Adversarial Dataset} featuring conflicting linguistic and acoustic cues. We have two setups. In the first setup, we inject a speech signal (generated by CosyVoice TTS~\cite{du2024cosyvoice}) that gives utters a wrong answer multiple times. In the second setup we inject a speech signal that utters the correct answer multiple times. In both setups the power of the adversarial speech is comparable or lower than the input audio. We intend to measure whether the model's rationale remains grounded in the audio (which is required to answer the question) or is misled by the text transcription.

\noindent \textbf{Results:} In Figure \ref{fig:adv_results}, we observe that the adversarial injections in fact result in very significant performance reductions when a wrong answer is injected in the audio. We observe for instance, that for the Sakura-Animal dataset, the performance drops from $\mathord{\sim}75\%$ to $\mathord{\sim}25\%$ for AF3. QWEN seems to be slightly more robust, but it still suffers from significant performance alterations because of the adversarial interventions. These results suggest that LALMs may rely more heavily on linguistic cues than on the underlying acoustic signal.

%\subsection{Grounded Masking}

\vspace{-.3cm}
\section{CoT Interventions}
\label{sec:cot_interventions}

In this study, we evaluate the faithfulness of a model's generated CoT to its final prediction (Q4: CoT-Output Faithfulness). By systematically modifying the content of the CoT, we test whether the reasoning process actually influences the model's output. Figure \ref{fig:pipeline_cot_intervention} illustrates this evaluation pipeline. The model is provided an input audio sample and a text prompt requiring step-by-step reasoning. Once the initial CoT is generated, we intervene by modifying the reasoning string before prompting the model to produce a final answer based on the altered CoT. Following \cite{lanham2023measuringfaithfulnesschainofthoughtreasoning}, we implement four intervention strategies:

\begin{itemize}
    \item \textbf{Filler Tokens:} To test if performance gains stem from \textit{extra test-time computation} rather than logic, we replace 0--100\% of CoT tokens with "lorem ipsum" dummy text in 5\% increments. This maintains the computational overhead (token count) while removing semantic information. High consistency under this intervention suggests the model relies on the additional processing time rather than the reasoning itself.

    \item \textbf{Early Answering:} To detect \textit{post-hoc reasoning}, we progressively truncate the CoT from the end. If the model maintains its prediction despite the loss of terminal reasoning steps, the decision was likely made prior to generating the full CoT, causing the reasoning to be a post-hoc justification.

    \item \textbf{Paraphrasing :} To identify \textit{encoded reasoning}, we use an \href{https://huggingface.co/unsloth/Mistral-Small-3.2-24B-Instruct-2506}{external LLM} to rewrite the CoT while preserving semantic meaning. By rephrasing text and punctuation, we eliminate potential hidden signals. A drop in consistency after paraphrasing would suggest model relies on information encoded in the phrasing rather than human-understandable logic.

    \item \textbf{Adding Mistakes :} We incrementally introduce logical errors using the same unbiased external LLM. The model is then prompted to continue its reasoning from the corrupted step. If the final prediction remains unchanged despite following a flawed chain, the model is unfaithful to its own stated reasoning process.
\end{itemize}

\noindent \textbf{Setup:} Consistent with our audio intervention experiments, we utilize all four tracks of SAKURA, along with the MMAR and MMAU datasets. We compare the final prediction generated from the modified CoT against the baseline answer (the output from the original CoT). Faithfulness is measured by the degree of consistency between these outputs.

\noindent \textbf{Results:} As shown in Figure \ref{fig:cot_intervention_grid}, across all intervention types, both AF3 and Qwen2.5-Omni demonstrate high faithfulness to their generated reasoning. Paraphrasing results show near-perfect consistency, indicating the models rely on semantic logic rather than specific linguistic encoding. In contrast, consistency scales directly with the integrity of the CoT: it drops significantly when reasoning is truncated (Early Answering), corrupted with logical errors (Adding Mistakes), or replaced with meaningless text (Filler Tokens). The decline in consistency suggests that the semantic integrity of CoT matters for the model output.

\vspace{-.4cm}
\section{Conclusions}
% In this paper, we investigated the faithfulness of CoT representations of LALMs with respect to the input audio, and with respect to audio. We have identified three criteria for faithful CoT with respect to input audio; Hallucination-free listening, holistic Listening, and Attentive Listening. We observe that both AF3, and QWEN exhibits hallucinatory CoTs depending on the dataset, measured by the heavy noise (-20dB SNR) and intense masking (100\% masking). We have also observed that the models remain brittle to adversarial injections, which suggests that the model's attention can easily get distracted from the input audio on which the model's being asked about. On the CoT interventions, we have observed that the CoTs produced by LALMs do not obviously exhibit posthoc reasoning, extra test-time computation or encoded reasoning effects. As mentioned however, our investigation reveals interesting shortcomings regarding faithfulness of the CoTs with respect to the input audio. We believe that future research can investigate methodologies to address these shortcomings.   
%hallucination-free listening, holistic listening, and attentive listening. 
In this work, we study the faithfulness of CoT reasoning in LALMs with respect to both the input audio and the final prediction. We defined three criteria for audio-faithful reasoning, and through our experiments show that both AF3 and Qwen2.5-Omni fail to consistently satisfy these properties. The models often hallucinate CoTs even when audio input is corrupted and are vulnerable to adversarial injections. 
At the same time, textual CoT interventions indicate that the generated reasoning is generally faithful to the model’s final prediction. Overall, our findings reveal a multimodal disconnect: LALMs are internally consistent with their text outputs but frequently fail to remain strongly grounded in the audio input, highlighting the need for stronger audio–text grounding mechanisms.

%as their attention is easily hijacked by irrelevant spoken transcripts rather than remaining focused on the target acoustic features they are asked to analyze. 
\section{Acknowledgments}
We acknowledge the support of the Natural Sciences and Engineering Research Council of Canada (NSERC) and the Digital Research Alliance of Canada (alliancecan.ca). Cem Subakan is supported by the NSERC project RGPIN-2023-05759. Mirco Ravanelli is supported by research funding, computational resources, and donations from NSERC, the Digital Research Alliance of Canada (alliancecan.ca), the Translated Imminent Program, and an Apple Seed Grant.

\section{Generative AI Use Disclosure}
Generative AI Use Disclosure: Large Language Models (LLMs) were used solely to assist with minor language editing, grammar correction, and polishing of the manuscript text. No AI tools were used to generate scientific ideas, experimental results, or analyses. All authors reviewed and verified the final content and take full responsibility for the accuracy and integrity of the work.

\bibliographystyle{IEEEtran}
\bibliography{mybib}

\end{document}